\pdfoutput=1

\documentclass{article}

\usepackage{microtype}
\usepackage{graphicx}
\usepackage{booktabs} 
\usepackage{caption}
\usepackage[caption=false]{subfig}

\usepackage[utf8]{inputenc} 
\usepackage[T1]{fontenc}    

\usepackage{hyperref}
\hypersetup{colorlinks,linkcolor=red}

\usepackage{url}            
\usepackage{amsfonts}       
\usepackage{nicefrac}       
\usepackage{microtype}      
\usepackage{todonotes}
\usepackage{amsmath}
\usepackage{boldline}
\usepackage{wrapfig}
\usepackage{mathtools}
\usepackage{threeparttable}
\usepackage{chngcntr}
\usepackage{makecell}
\usepackage{xspace}
\usepackage{balance}

\usepackage{multibib}

\usepackage{subfiles}

\newcommand\mypara[1]{\vspace{0.5mm}\noindent\textbf{#1}\xspace}

\usepackage{hyperref}

\usepackage[accepted]{icml2021}

\newcommand{\Skip}[1]{{}}
\newcommand{\tocite}[1]{{}}

\newcommand{\ie}{\textit{i}.\textit{e}.\ }
\newcommand{\eg}{\textit{e}.\textit{g}.\ }
\newcommand{\etc}{\textit{etc}.\ }
\newcommand{\myfig}[1]{Figure~\ref{#1}}
\newcommand{\mytable}[1]{Table~\ref{#1}}

\newcommand{\csection}[1]{
    \section{#1}
}

\newcommand{\csubsection}[1]{
    \subsection{#1}
}

\icmltitlerunning{Megaverse: Simulating Embodied Agents at One Million Experiences per Second}

\newcites{appendix}{References}

\begin{document}

\twocolumn[{

\icmltitle{Megaverse: Simulating Embodied Agents at \\ One Million Experiences per Second}



\icmlsetsymbol{equal}{*}

\begin{icmlauthorlist}
\icmlauthor{Aleksei Petrenko}{Intel,USC}
\icmlauthor{Erik Wijmans}{Intel,GT}
\icmlauthor{Brennan Shacklett}{Stanford}
\icmlauthor{Vladlen Koltun}{Intel}
\end{icmlauthorlist}

\icmlaffiliation{USC}{University of Southern California}
\icmlaffiliation{Intel}{Intel Labs}
\icmlaffiliation{GT}{Georgia Institute of Technology}
\icmlaffiliation{Stanford}{Stanford University}

\icmlcorrespondingauthor{Aleksei Petrenko}{petrenko@usc.edu}

\icmlkeywords{Machine Learning, Reinforcement Learning, Simulation, ICML}

{\centering
\vspace{0.1in}
\newlength{\colw} 
\setlength{\colw}{0.33\linewidth}
\setlength{\tabcolsep}{1.0mm} 
\begin{tabular*}{1.0\linewidth}{ m{\colw} m{\colw} m{\colw}}
  \includegraphics[width=\colw]{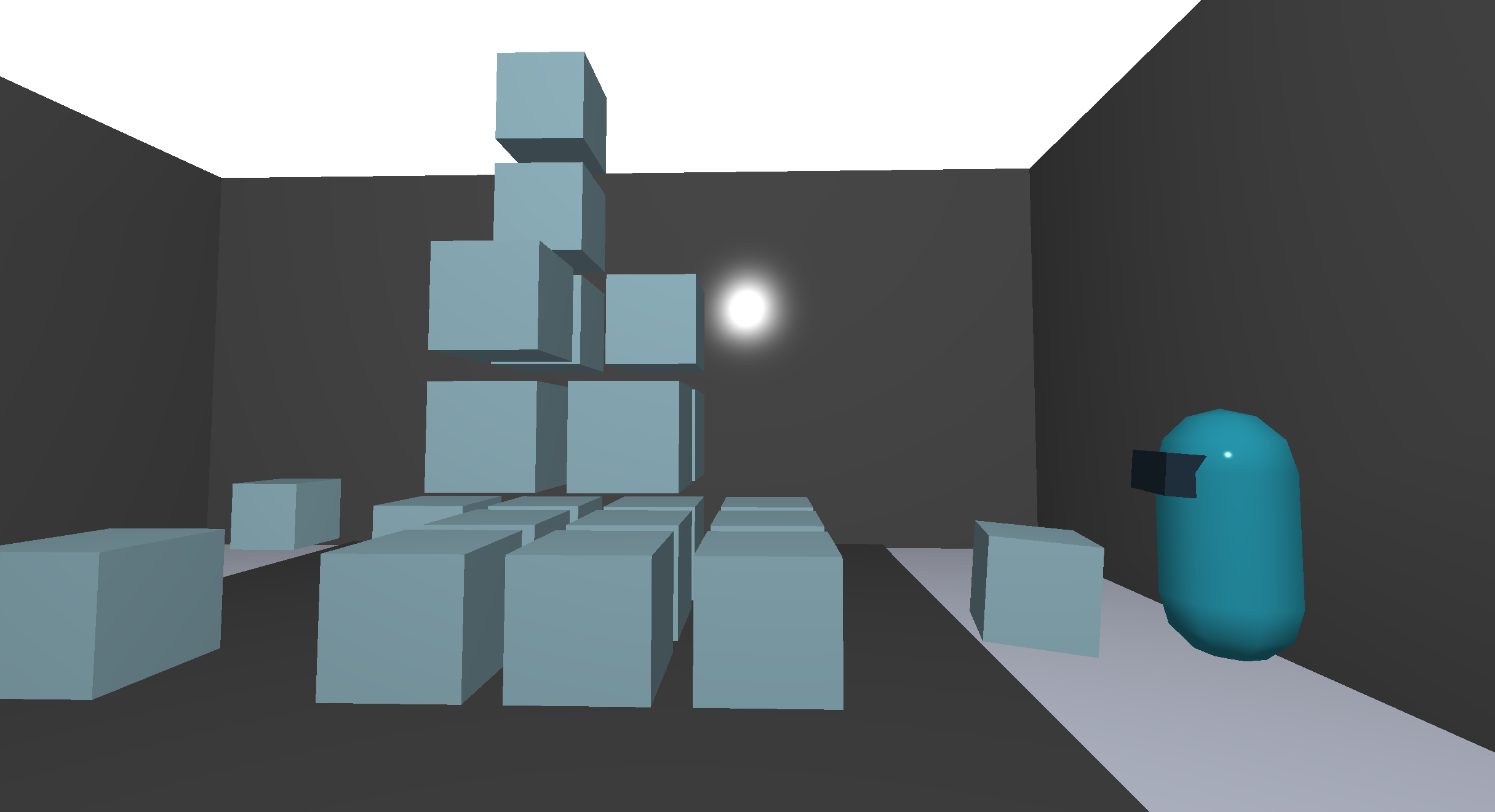} &
  \includegraphics[width=\colw]{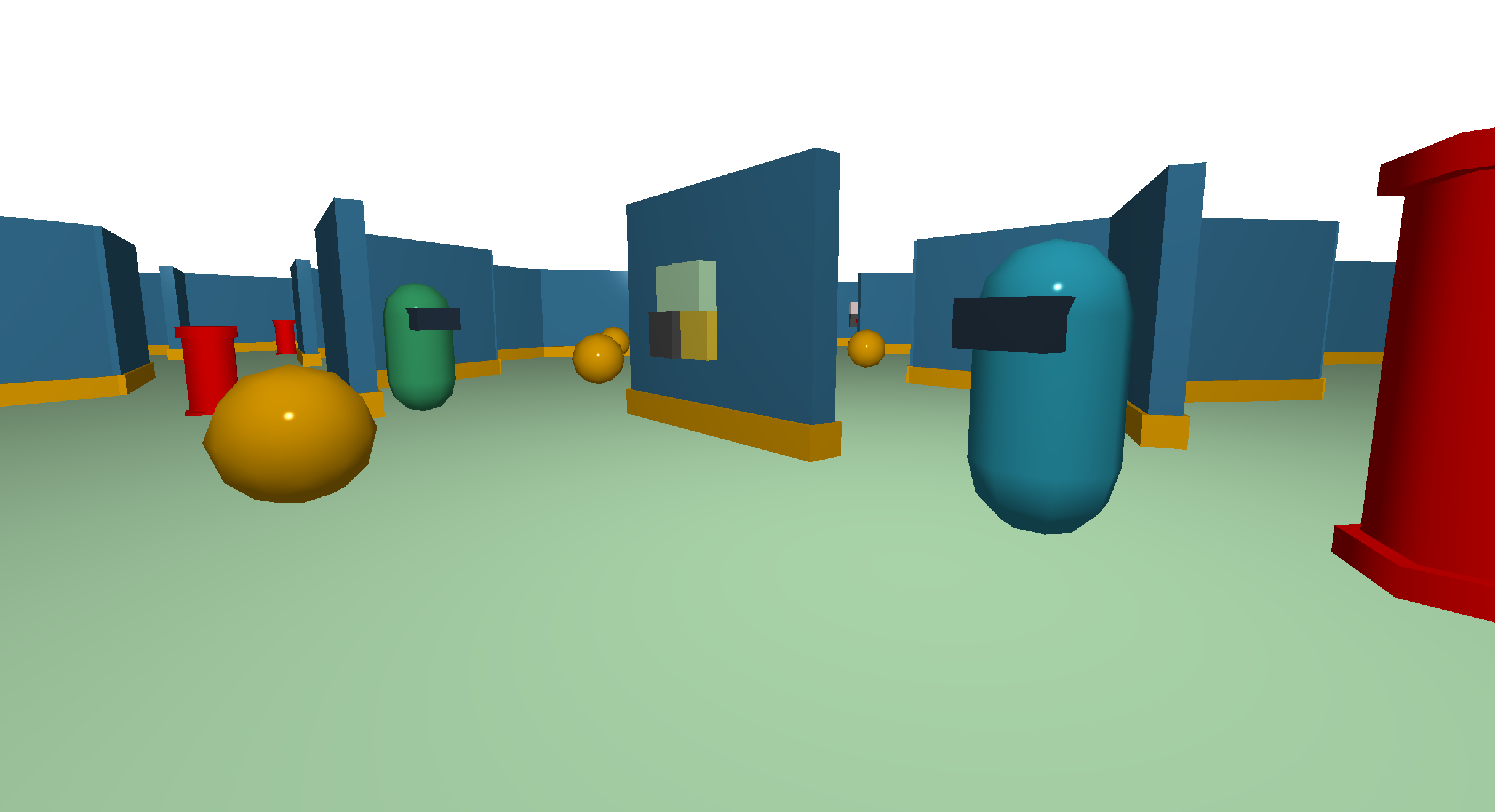} &
  \includegraphics[width=\colw]{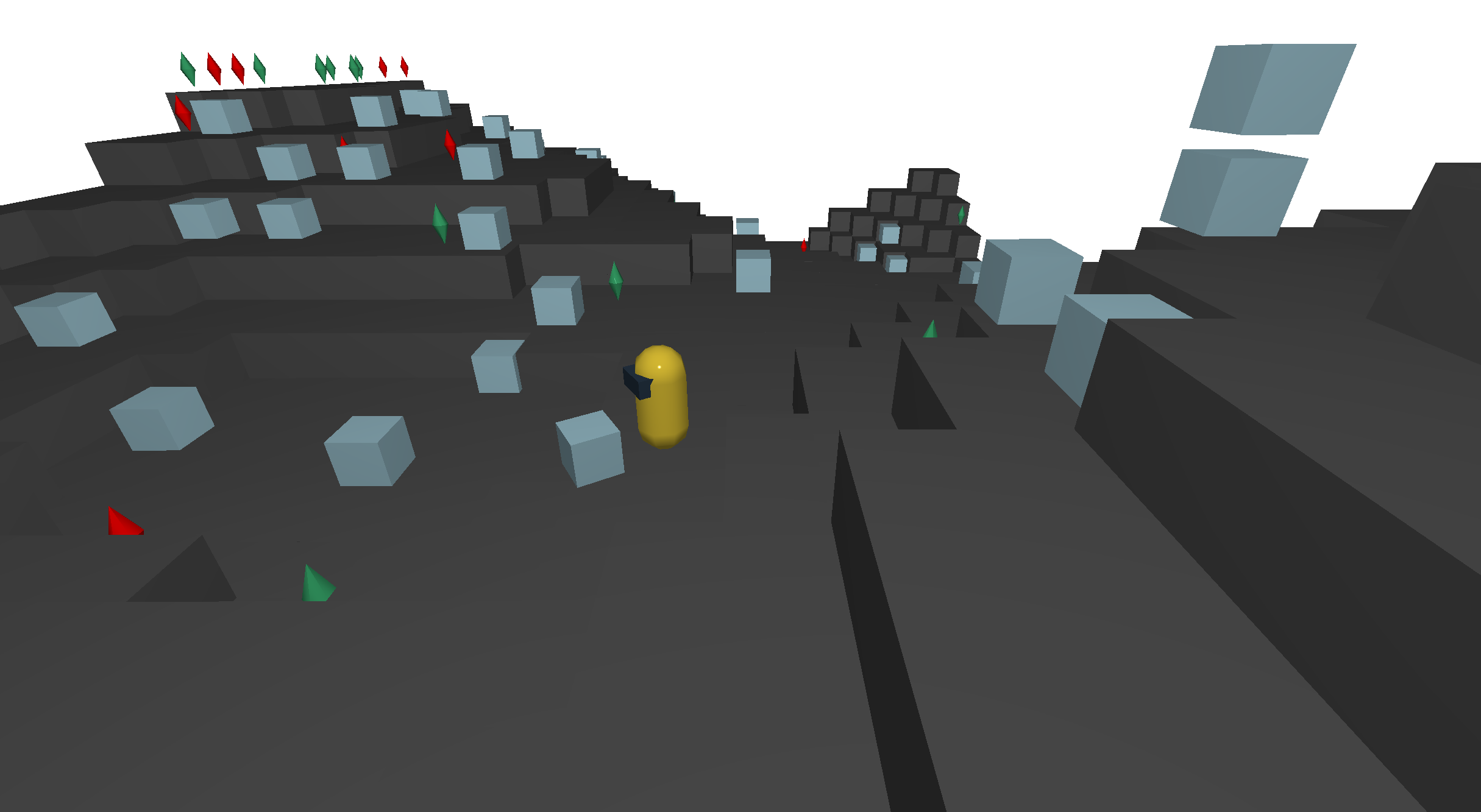}
\end{tabular*}
\setlength{\colw}{0.495\linewidth}%
\begin{tabular*}{1.0\linewidth}{ m{\colw} m{\colw} }
    \includegraphics[width=\colw]{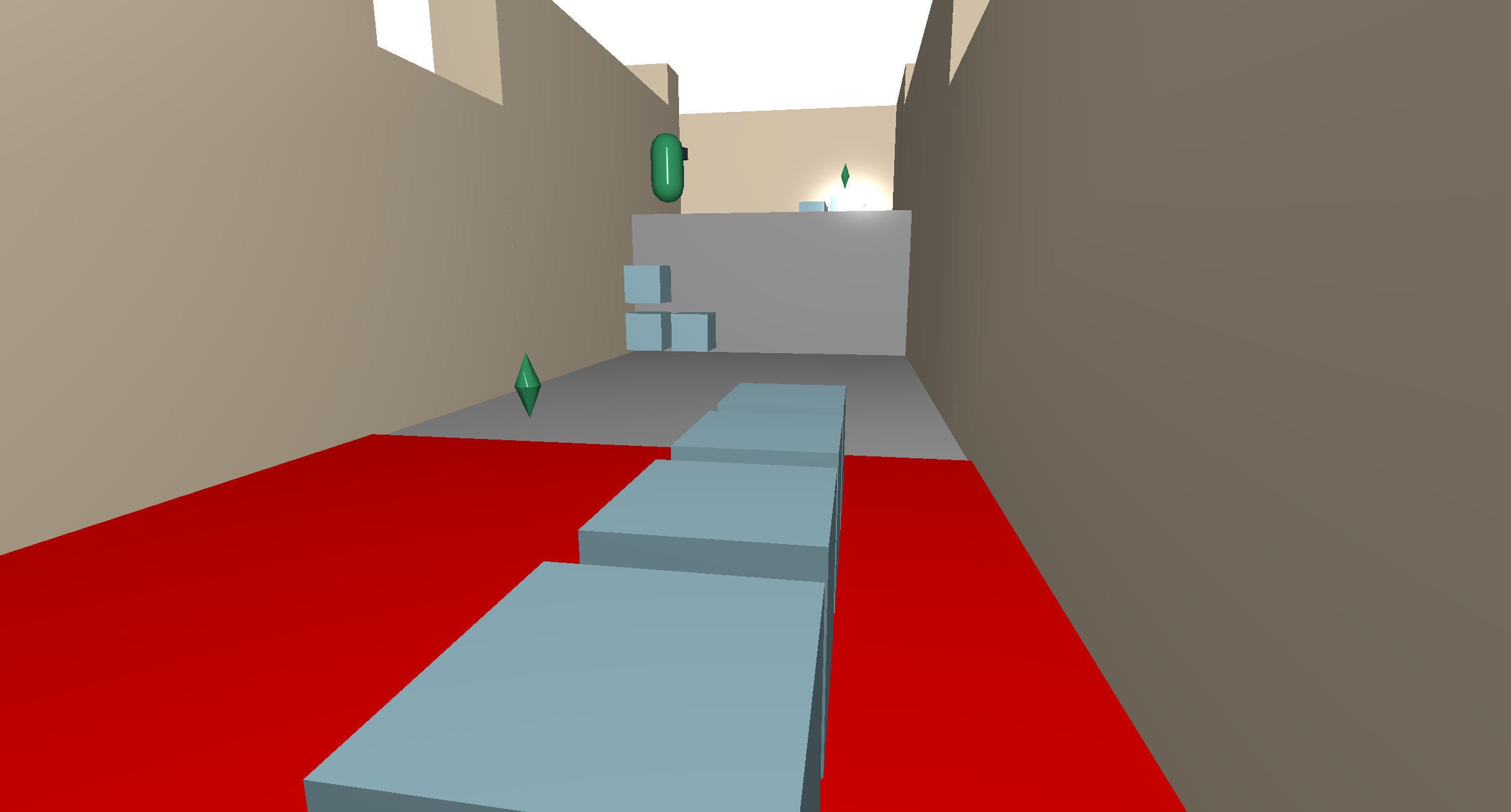} &
    \includegraphics[width=\colw]{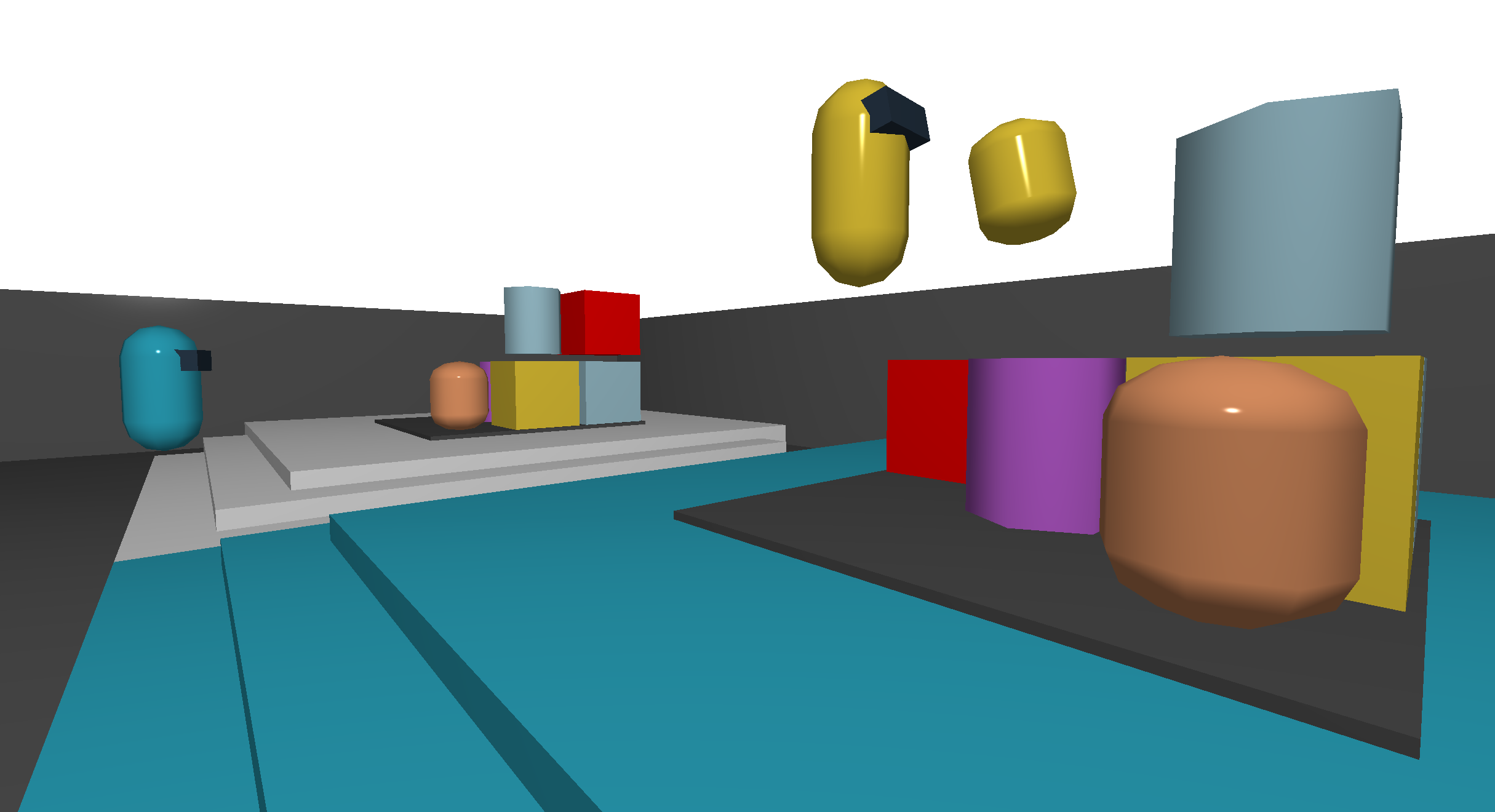}
\end{tabular*}%
}  \\[-10pt]
  \label{fig:teaser}
  \captionof{figure}{Megaverse is a new research platform that supports simulating embodied agents in immersive interactive 3D environments at over 1,000,000 experiences per second on a single 8-GPU server. Video demonstrations of Megaverse scenarios and behaviors of trained agents are available at \url{www.megaverse.info}}

\vskip 0.3in
}]



\printAffiliationsAndNotice{}  

\begin{abstract}
We present Megaverse, a new 3D simulation platform for reinforcement learning and embodied AI research. The efficient design of our engine enables physics-based simulation with high-dimensional egocentric observations at more than 1,000,000 actions per second on a single 8-GPU node. Megaverse is up to 70x faster than DeepMind Lab in fully-shaded 3D scenes with interactive objects. We achieve this high simulation performance by leveraging batched simulation, thereby taking full advantage of the massive parallelism of modern GPUs. We use Megaverse to build a new benchmark that consists of several single-agent and multi-agent tasks covering a variety of cognitive challenges. We evaluate model-free RL on this benchmark to provide baselines and facilitate future research.
The source code is available at \url{www.megaverse.info}
\end{abstract}

\csection{Introduction}
\label{intro}

Since the advent of deep learning, progress in the field has been driven by two factors: the availability of large amounts of data and the ability to train big and expressive parametric models on this data. Deep reinforcement learning is no exception: the existence of fast, efficient, and challenging simulation environments that can generate millions of trajectories allowed researchers to quickly iterate, experiment with learning at scale, and ultimately develop improved models and learning algorithms.

The majority of RL environments used in contemporary Deep RL research with high-dimensional observations are based on repurposed game engines.
For example, the Arcade Learning Environment (ALE) \cite{ale}, VizDoom \cite{vizdoom}, and DeepMind Lab \cite{dmlab} use the engines of Atari 2600, Doom, and Quake III, respectively. While reusing existing graphics pipelines reduces development costs, it can be a suboptimal basis for RL environments.

Game engines have been designed and optimized to render a single complex scene at resolutions and framerates tuned for human perception.
The simulation of game physics proceeds in very small steps such that the resulting motion and animations appear smooth when the end user interacts with the game world in real time. Environments for RL research, however, have drastically different requirements. State-of-the-art learning algorithms typically consume large amounts of low-resolution observations rendered much faster than real time. Turning a game engine into a high-throughput simulator for RL thus requires using OS-level parallelism, where individual game instances are run at the same time in parallel processes. This prevents the engines from efficiently sharing resources, increases memory consumption, and ultimately fails to fully utilize the throughput of modern hardware accelerators. In addition, due to the high granularity of physics simulation, researchers often use techniques like frame-skipping \cite{ale} to simplify credit assignment, thereby wasting computation on synthesizing views that are never observed by agents.

We argue that a purpose-built rendering and simulation engine for RL research can eliminate inefficiencies of the existing environments and therefore accelerate research in reinforcement learning and artificial intelligence. We present Megaverse, a new platform for embodied AI. Built from scratch as a lightweight simulator, it leverages batched rendering~\cite{shacklett2021bps} to fully utilize the throughput of modern GPUs and render up to $1.1 \times 10^6$ observations per second on a single 8-GPU node at $128 \times 72 \times 3$ resolution. By default we simulate physics at lower frequency compared to traditional 3D engines, which eliminates the need for frame-skipping and thus enables an experience collection rate up to $\sim$20x faster than Atari and $\sim$70x faster than DeepMind Lab on hardware commonly found in deep learning research labs.

Since Megaverse was designed to render an arbitrary number of viewpoints and scenes in parallel, it naturally supports multi-agent simulation. While traditional RL engines either do not support multi-agent training \cite{ale} or require a slow network setup to enable it \cite{vizdoom}, Megaverse can simulate the experiences of dozens of agents in the same environment without loss of performance, in both collaborative and competitive (self-play) settings.

We introduce Megaverse as a simulation platform that can be used to create virtual worlds for deep RL and embodied AI research. We release eight environments built on top of Megaverse that cover an array of embodied cognitive tasks and prove to be hard for modern RL algorithms. Our benchmark, \emph{Megaverse-8}, addresses challenges such as navigation, exploration, and memory. All of the environments are procedurally generated, and can therefore be used to investigate generalization of trained agents~\cite{cobbe2019procgen}. Many of the challenges require our agents to learn nontrivial physics-based environment manipulation, which has previously been a feature of resource-intensive high-fidelity simulators~\cite{ai2thor}.

A key goal of Megaverse is to democratize deep RL research. State-of-the-art results in RL have primarily been a prerogative of large research labs with access to vast computational resources. A fast and efficient simulation engine that supports interactive immersive environments that call for advanced embodied cognition can enable rapid community-wide experimentation and iteration, thus accelerating progress in the field.

\csection{Prior work}
\label{prior_work}

The first artificial agents were typically confined to miniature grid worlds or board games \cite{tesauro1995tdgammon}. These relatively simple environments were nevertheless a challenge for early intelligent systems. They allowed researchers to hone the foundations of reinforcement learning theory and general-purpose learning algorithms \cite{sutton1992reinforcement}.

With the advent of powerful function approximators, researchers turned their attention to 2D computer games as a new challenge for artificial agents. The DQN algorithm \cite{mnih-dqn-2015} demonstrated the ability to learn directly from high-dimensional pixel observations, matching or exceeding human-level performance on multiple Atari games.

Rapid progress on Atari-like benchmarks led to a phase transition: a new generation of AI research platforms brought \emph{immersive} simulators. In contrast to the flatland of arcade games, the real world is immersive and 3-dimensional. In order to successfully operate in the real world, artificial agents have to master skills such as spatiotemporal reasoning and object manipulation. Simulators derived from first-person 3-dimensional video games, such as VizDoom \cite{vizdoom}, DeepMind Lab \cite{dmlab}, and MineRL \cite{guss2019minerl} were among the first to offer virtual embodiment and egocentric perception.

Immersive simulators vary in their fidelity and throughput. Advanced simulation platforms built on top of modern 3D engines, such as Unity \cite{ai2thor} or Unreal \cite{dosovitskiy17carla}, trade simulation speed for high-fidelity graphics. These are useful in studying sim-to-real transfer and perceptual aspects of embodied AI, but end-to-end learning of non-trivial skills and behaviors in these environments requires massive computational resources \cite{ddppo}.

Other efforts focus on increasing the behavioral complexity of simulated worlds. Interactions arising in such environments allow researchers to study non-trivial behaviors and strategies learned by artificial agents.
Environments such as Dota 2 \cite{openai2019dota} and StarCraft II \cite{dmstarcraft2} are complex modern strategy games, and playing these games at a human level requires advanced long-term planning. However, these games are not fully immersive, they provide only a structured top-down view of the environment.
The OpenAI Hide-and-Seek project \cite{hide-n-seek} investigates sophisticated behaviors emerging in an environment with full physics simulation. Unlike the strategy games mentioned above, this environment simulates egocentric perception, although only 1D Lidar-like sensing is supported.

Capture the Flag \cite{dmquakescience}, which is based on the Quake III engine, demonstrated great potential of reinforcement learning in immersive environments, but the project relied on complex closed-source compute infrastructure. With Megaverse we aim to build a platform that enables the exploration of advanced embodied cognition at or beyond the level of Capture the Flag and Hide-and-Seek, with full physics simulation and high-dimensional pixel observations, without requiring tens or hundreds of servers for experience collection.

Megaverse is not the first initiative that aims at building an RL simulator from scratch. For example, MINOS \cite{savva2017minos} and its successor Habitat \cite{habitat19iccv} use a purpose-built rendering engine and support high-resolution textured scenes based on 3D scans of real environments \cite{ChangDFHNSSZZ17matterport}.  These and other environments can be used as testbeds for embodied challenges such as navigation \cite{Anderson2018} and rearrangement \cite{batra2020rearrangement}. Most related to our work is \citet{shacklett2021bps} who demonstrate that considerable speedups can be gained via batched simulation, synthesizing the observations in many environments simultaneously. Whereas \citet{shacklett2021bps} only examined very simplistic physics (simple collisions with static geometry), with a single agent per environment, we apply batched simulation to multi-agent environments with complex interactivity.

The idea of batched GPU-accelerated simulation has also been applied in the context of continuous control such as robot locomotion and dexterous manipulation \cite{liang2018pre-isaac-gym}. Perhaps a combination of batched rendering and batched physics can pave the way for a new generation of fast high-fidelity simulators that will become a core part of future AI research infrastructure.

\csection{Megaverse}
\label{megaverse}

Megaverse is a purpose-built simulation platform for embodied AI research. Our engine can simulate fully immersive 3D worlds with multiple agents interacting with each other and manipulating physical objects, at more than $10^6$ experiences per second on a single node. The agents perceive the world through high-dimensional observations, rendered with dynamic shading and simulated lighting.

We introduce a number of performance optimizations that are instrumental in unlocking this performance regime. Our discretized physics approach (Section \ref{discretized_physics}) allows us to streamline the computation of non-trivial physical interactions between simulated objects. The parallel architecture built around the batched Vulkan renderer \cite{shacklett2021bps} (Section \ref{large_batch_rendering}) enables the production of hundreds of observations in a single pass, drastically reducing the required amount of communication between hardware components. Another algorithmic optimization is 3D geometry simplification (Section \ref{geometry_optimization}). The following sections describe these architectural choices and optimizations in more detail.

\begin{figure*}[t]
  \centering
  {\small
   \includegraphics[width=0.9\textwidth]{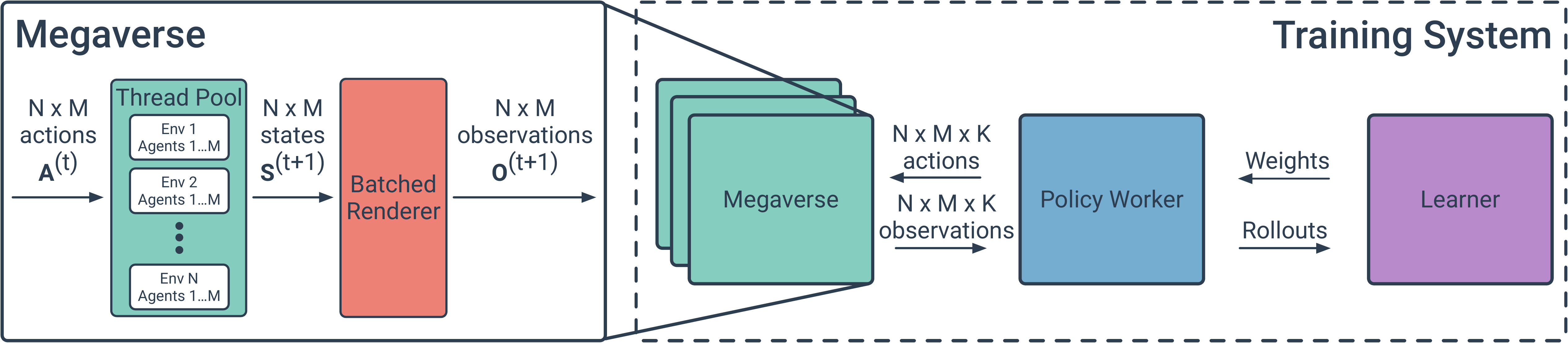}
   \vspace{-5pt}
   \caption{\textbf{Megaverse parallel architecture} in the context of a reinforcement learning system. Synchronous or asynchronous training sequentially interacts with $K \geq 1$ instances of Megaverse, each hosting $N\geq1$ environments with $M\geq1$ agents. Megaverse parallelizes physics computations on CPU, after which the entire vector of observations is rendered in a single pass.}
   \vspace{-5pt}
   \label{fig:architecture}
  }
\end{figure*}

\csubsection{Discretized continuous physics}
\label{discretized_physics}

Full simulation of physical contact and collisions between dozens of objects can be computationally expensive. Most physics engines require a small simulation step to avoid unrealistic interpenetration of objects caused by movement interpolation. Megaverse works around this by discretizing some of the physical interactions. Even though movement and collision checking for agents are fully continuous, we simplify the simulation of more complex interactions, such as object stacking. This enables non-trivial object manipulation, \eg building staircases and bridges using physical objects, or moving objects around while solving rearrangement tasks without prohibitively expensive high-frequency simulation of full contact forces.

Specifically, we use a voxel grid data structure: the agents are free to pick up objects anywhere and move them continuously, but placement of objects is only allowed at discretized locations in space, similar to MineRL \cite{guss2019minerl}. This has two major benefits. First, proximity checks and other spatial queries become trivial $\mathcal{O}(1)$ operations. More importantly, collision checking even with hundreds of objects also becomes extremely fast. We take advantage of caching mechanisms based on axis-aligned bounding boxes implemented in the Bullet physics engine \cite{coumans2019bullet}. Since most of the interactive objects reside in axis-aligned voxels, a simple check based on bounding box intersections eliminates the vast majority of potential collision candidates.

\csubsection{Large batch simulation and rasterization}
\label{large_batch_rendering}

One of the key design features of our implementation is an efficient parallel architecture (\myfig{fig:architecture}). A single instance of the traditional game-based RL environment can simulate only a single virtual world and render a single observation per simulation step. One instance of Megaverse can advance hundreds of environments, each containing multiple agents, in parallel. This allows our engine to take full advantage of the massive parallelism of modern computing platforms.

Formally, for environment $i$ our system maintains an internal state $\mathbf{s}_i^{(t)} = [s_{(i, 1)}^{(t)}, \ldots, s_{(i, N_i^{(t)})}^{(t)}]$, where ${s_{(i, n)}^{(t)} \in \text{Sim(3)}}$\footnote{The group of all rotations, translations, and scalings in 3D.} and $N_i^{(t)}$ is the number of entities in the environment $i$ at time step $t$ (\ie interactive objects, walls, agent bodies, cameras, etc.). Then, given a tensor of actions ${\mathbf{A}^{(t)} \in \mathcal{A}^{\text{NumEnvs} \times \text{NumAgents}}}$, where $\mathcal{A}$ is the set of all available actions, we update the internal state to produce ${\mathbf{S}^{(t+1)} = [\mathbf{s}_i^{(t + 1)}, \ldots, \mathbf{s}_{\text{NumEnvs}}^{(t + 1)}]}$. Taken together, the dynamics can be summarized as follows:
\begin{align}
    \mathbf{S}^{(t+1)} &= \text{Physics}(\mathbf{S}^{(t)}, \mathbf{A}^{(t)}) \\
    \mathbf{O}^{(t+1)} &= \text{Render}(\mathbf{S}^{(t+1)})
\end{align}
where $\mathbf{O}^{(t+1)}$ is the rendered observations while Physics and Render are \emph{batched} modules that are responsible for their own parallelization.  See \myfig{fig:architecture} for a visual depiction.

Physics simulation is parallelized on the CPU by scheduling state updates for individial environments on a thread pool with a configurable number of threads. In order to parallelize the rasterization step we adopt the optimized batched rendering pipeline proposed by \citet{shacklett2021bps}. This technique takes advantage of the fact that modern GPUs excel at rendering relatively small numbers of ultra-high-resolution images. The renderer bundles together all the rendering commands corresponding to individual agents and makes a single request to the GPU to render all the observations. This massively cuts down the required amount of communication between CPU and GPU, and helps the renderer achieve high GPU utilization. The impact of this technique on sampling performance is investigated in Section \ref{perf_ablation}.
For compatibility with existing training systems, by default we transfer the resulting rendered images to CPU memory, although it is also possible to expose them directly as PyTorch GPU-side tensors \cite{pytorch}.

\csubsection{3D geometry optimization}
\label{geometry_optimization}

\begin{figure}[htb]
    \centering
    \includegraphics[width=0.9\linewidth]{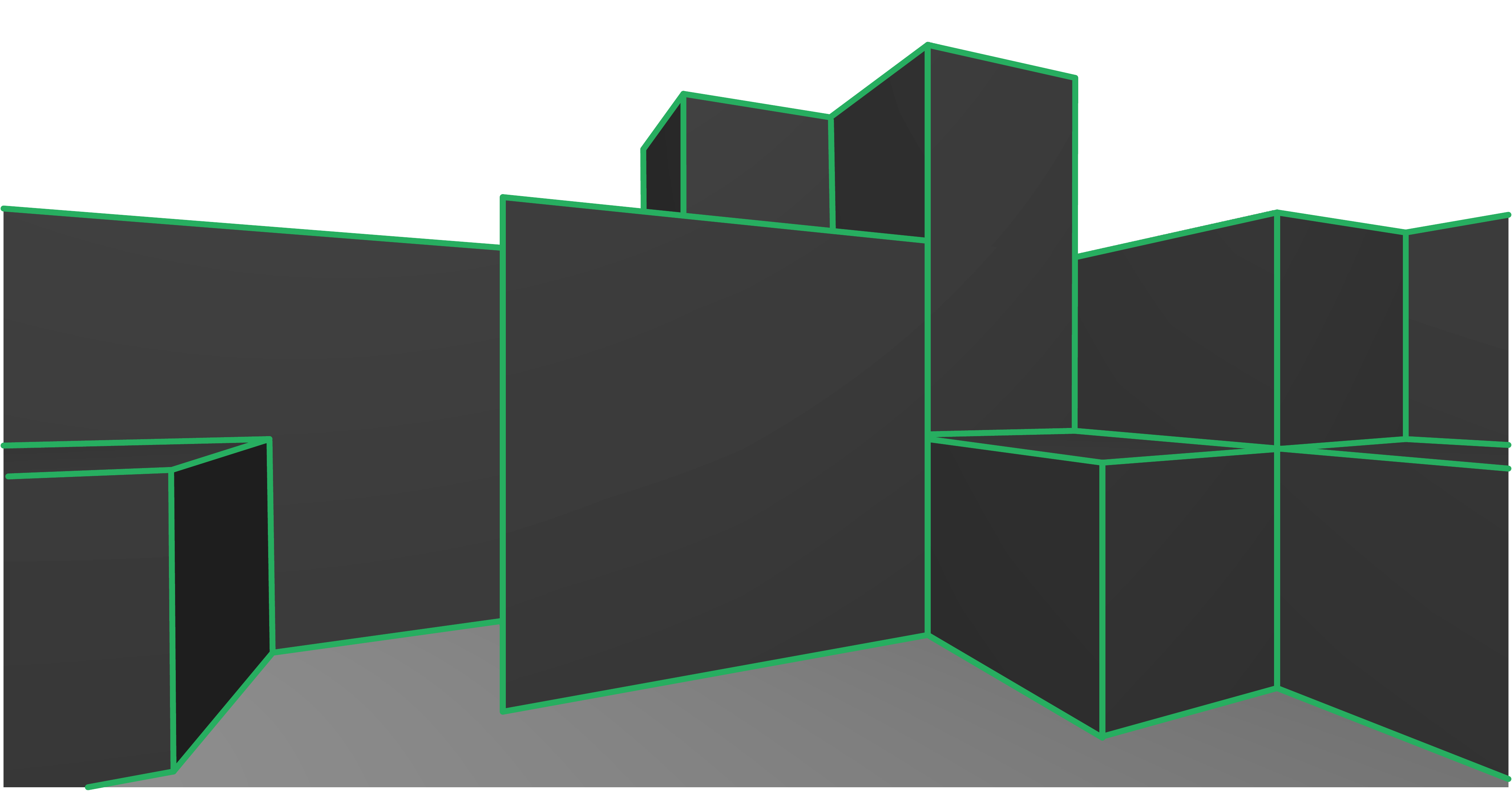}
    \caption{After procedurally generating the environment layout, we minimize the number of primitives by coalescing adjacent voxels where possible.}
    \label{fig:geometry_optimization}
    \vspace{-2pt}
\end{figure}

Voxelized geometry allows us to speed up physics calculations, but it is not the most efficient way to represent procedurally generated environment layouts, especially in scenarios with non-trivial 3D landscapes. A naive way to visualize such layouts would require rendering thousands of individual voxels that make up the environment. Instead, at the beginning of every episode, after the random landscape is generated, we merge adjacent voxels into a small number of enclosing parallelepipeds (Figure \ref{fig:geometry_optimization}). While finding the optimal solution for this problem is NP-hard, a greedy $\mathcal{O}(n)$ algorithm (where $n$ is a total number of non-empty voxels) is sufficient to significantly reduce the number of geometric primitives in the environment. We study the impact of this technique on sampling throughput in Section~\ref{sec:perf}.

\begin{figure*}
\centering
\newlength{\colww} 
\setlength{\colww}{0.48\linewidth}
\setlength{\tabcolsep}{0.1mm} 
\begin{tabular}{ m{\colww} m{\colww} m{\colww} m{\colww} }
    \subfloat[ObstaclesEasy]{\includegraphics[width=\colww]{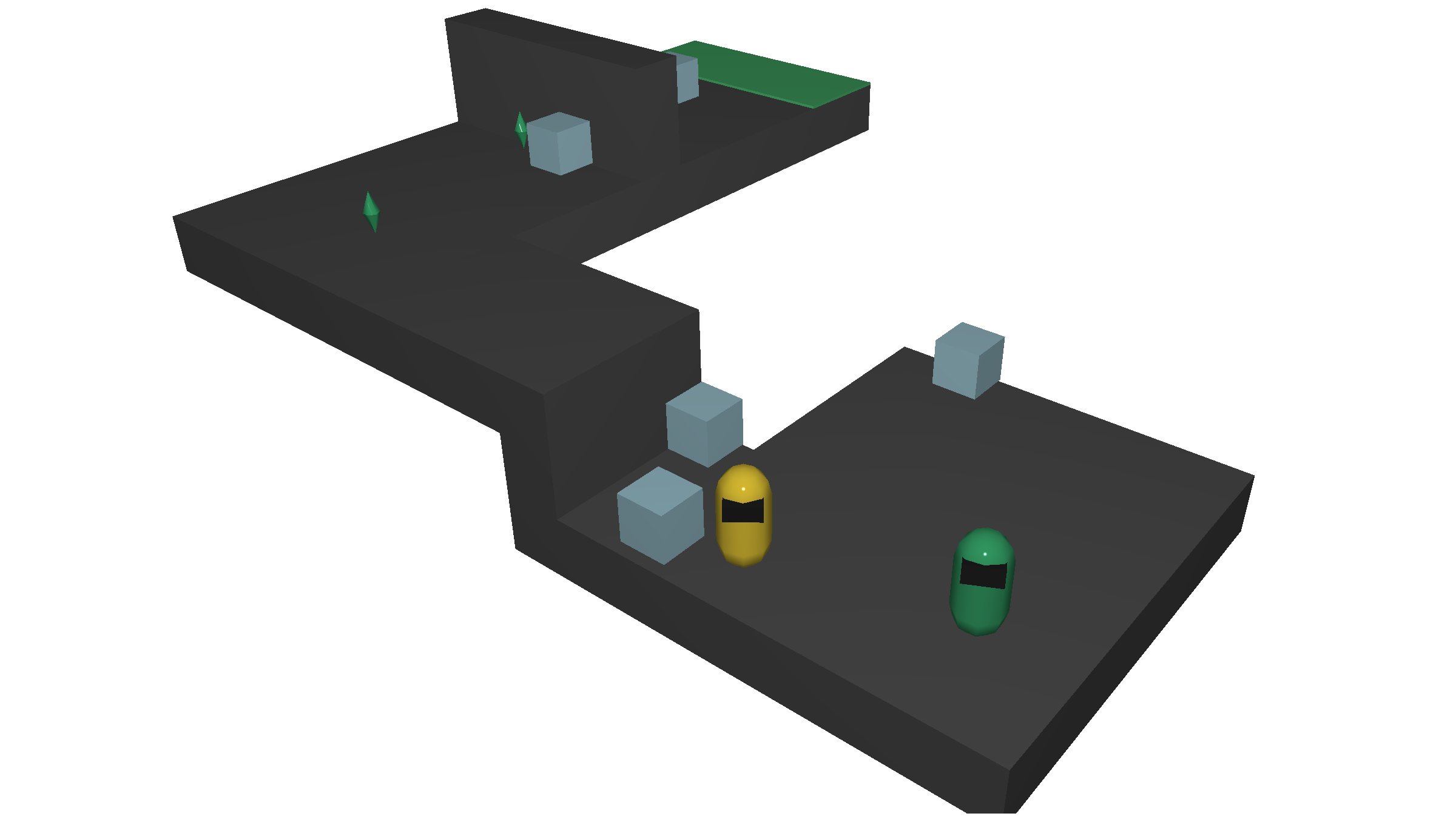}\label{fig:obseasy}}
    & \subfloat[ObstaclesHard]{\includegraphics[width=\colww]{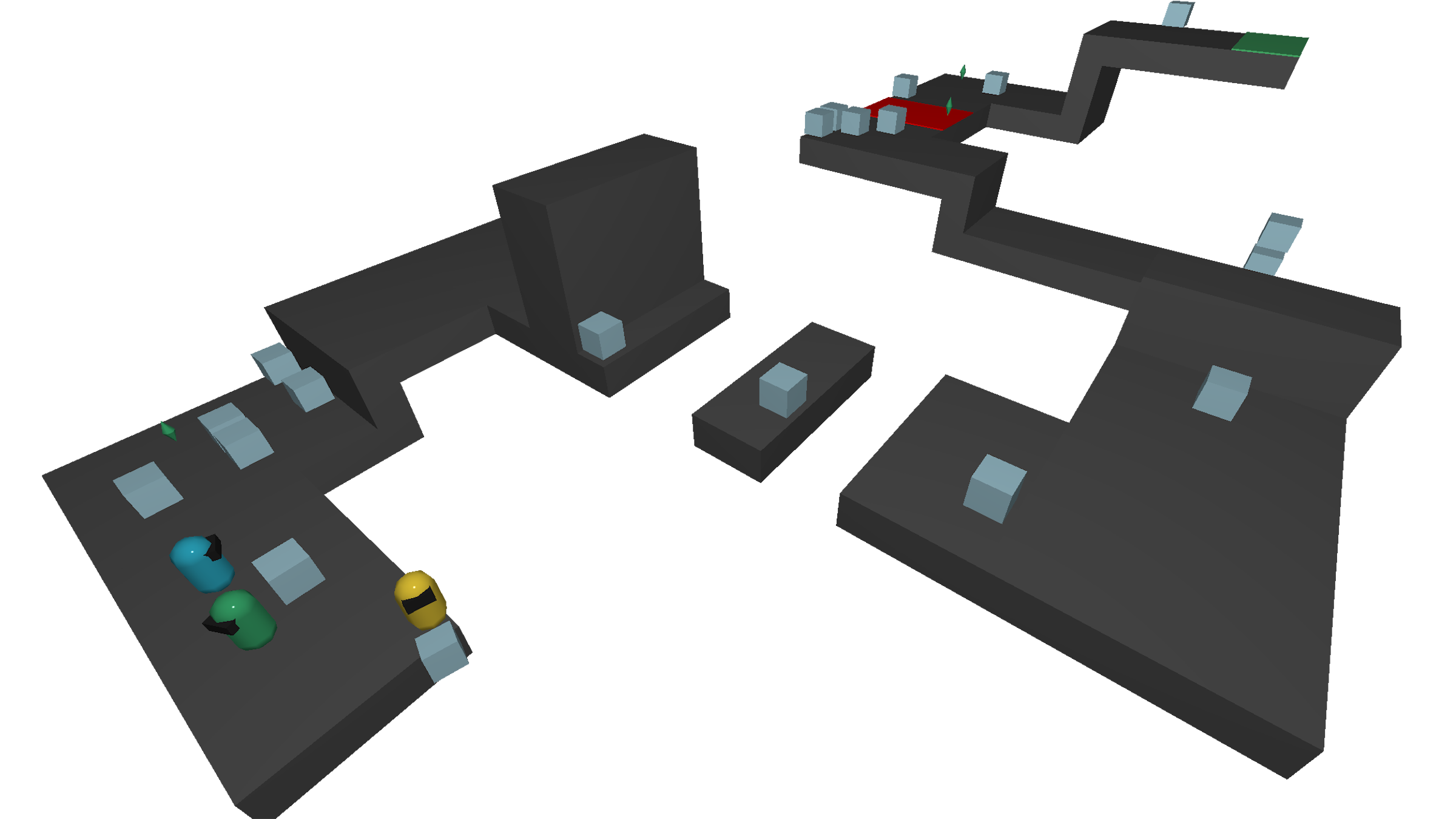}\label{fig:obshard}}
    \\
    \subfloat[Collect]{\includegraphics[width=\colww]{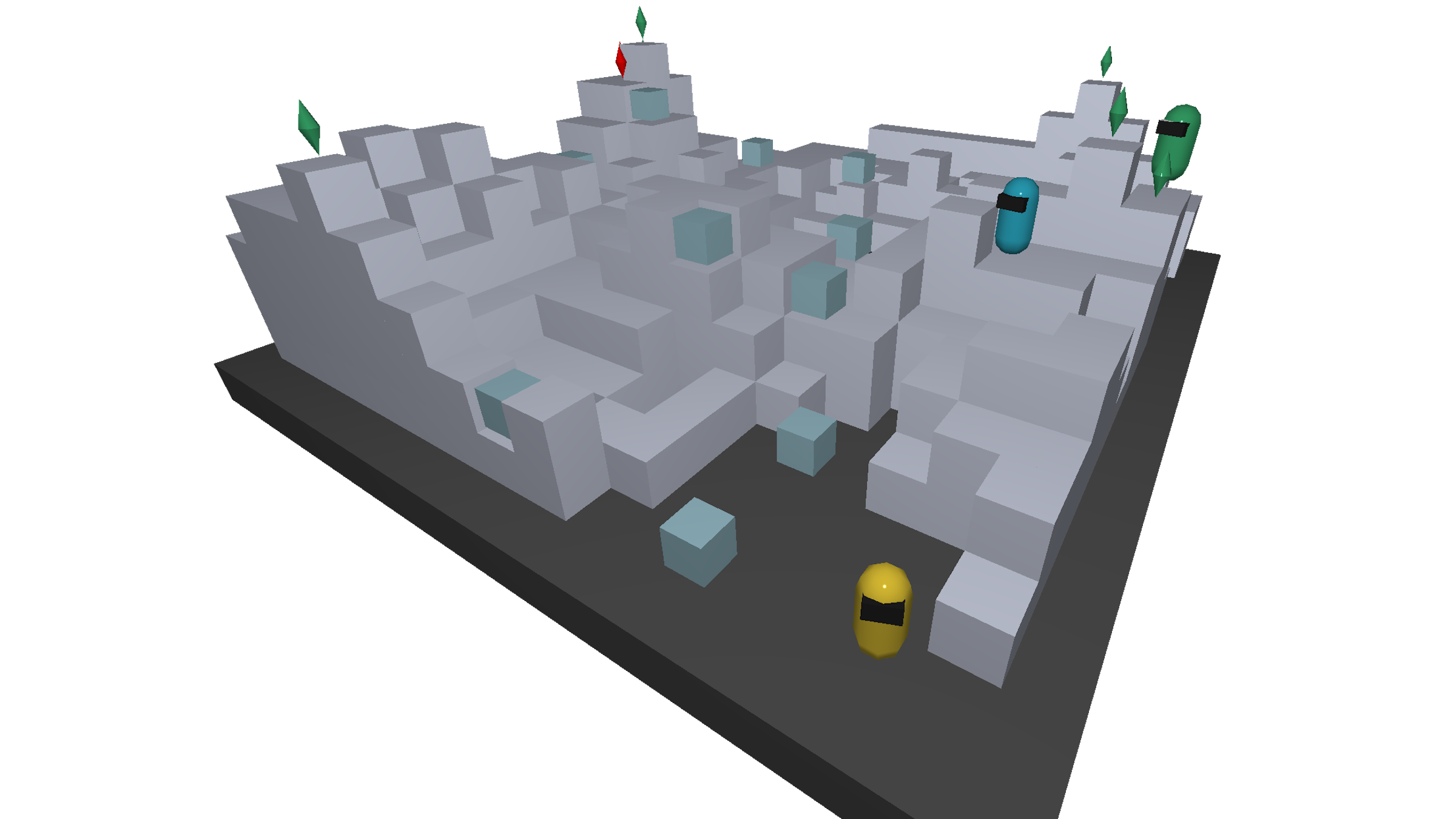}\label{fig:collect}}
    & \subfloat[Sokoban]{\includegraphics[width=\colww]{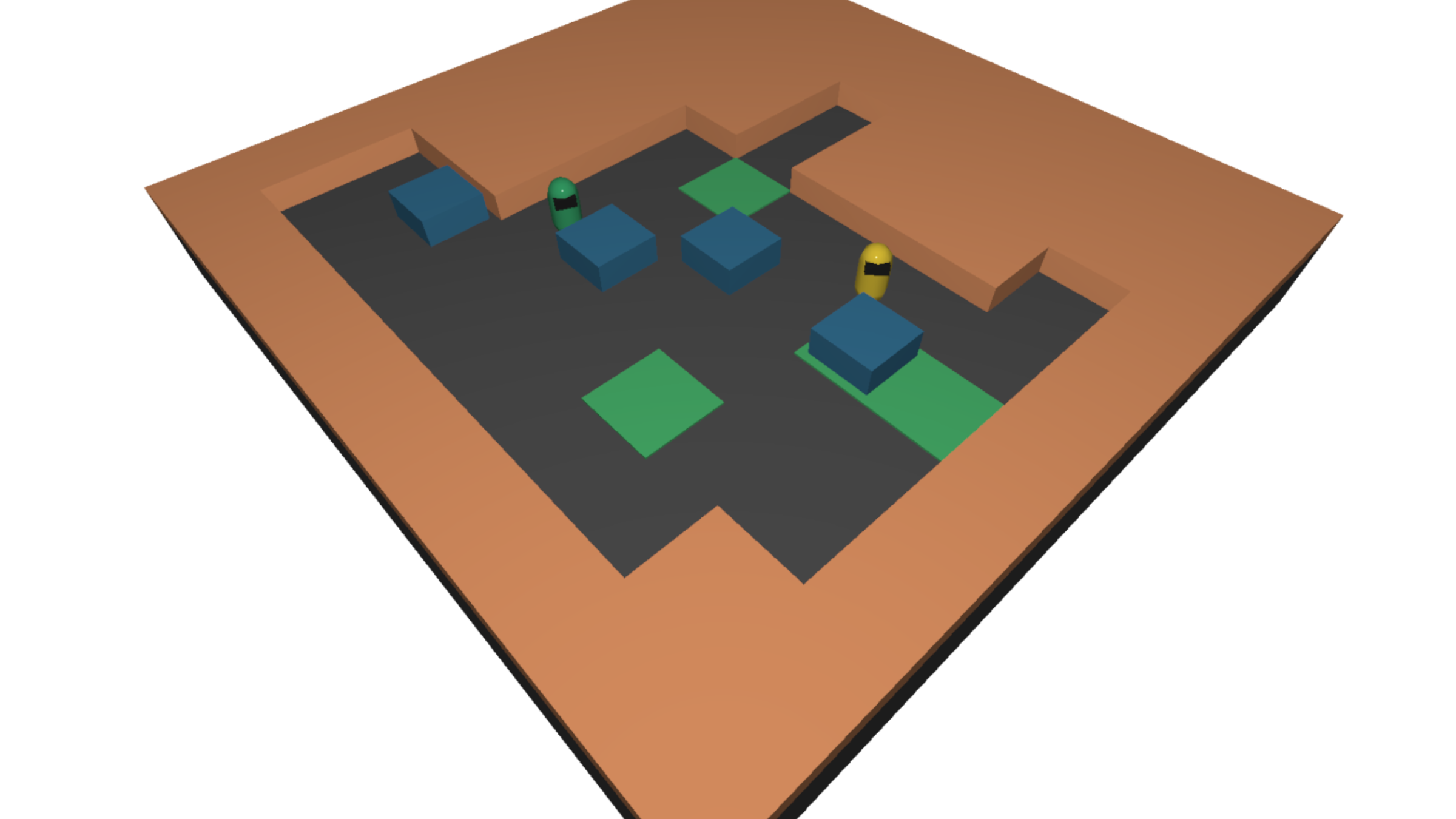}\label{fig:socoban}}
    \\
    \subfloat[HexExplore]{\includegraphics[width=\colww]{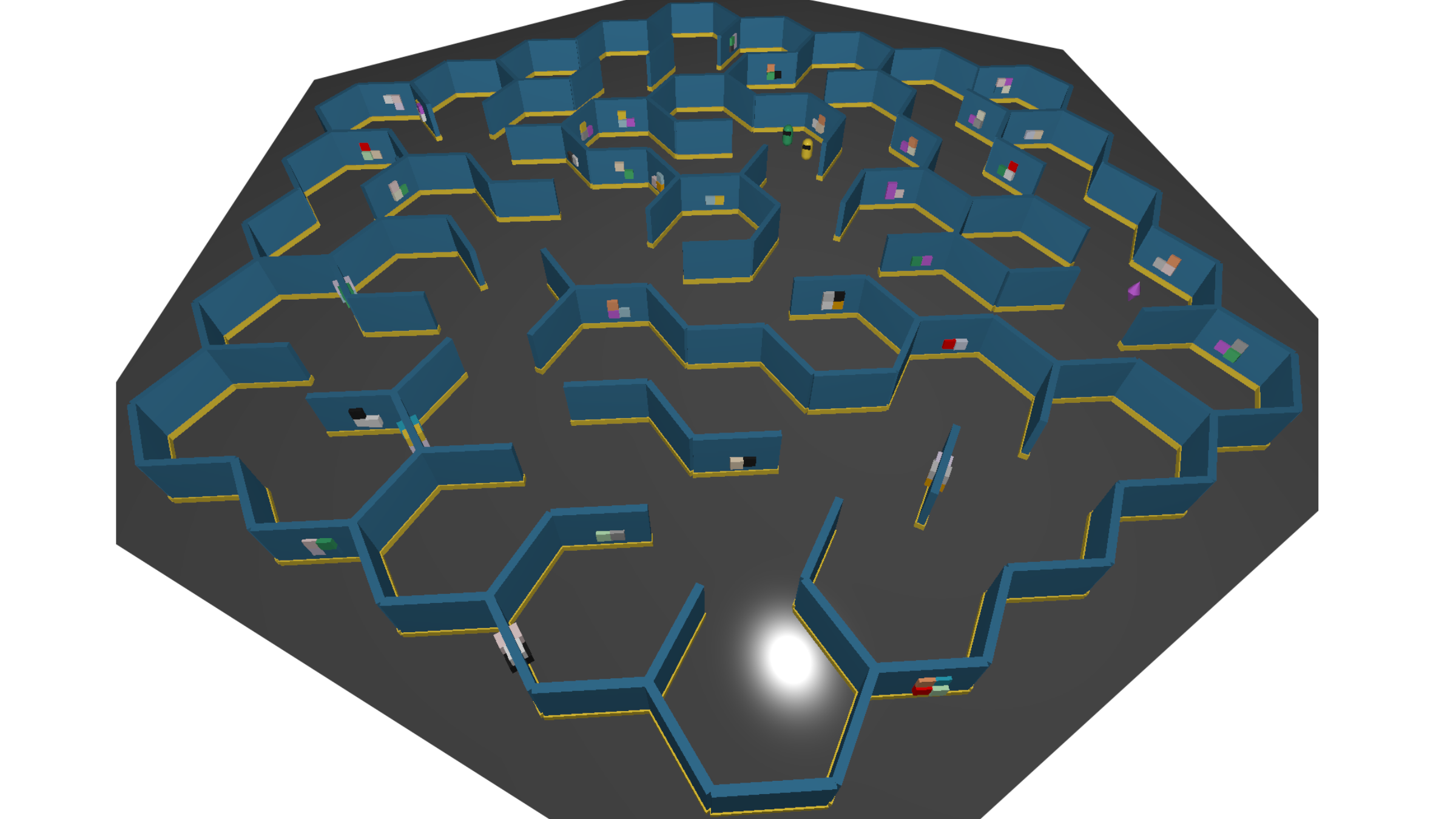}\label{fig:hexexplore}}
    & \subfloat[HexMemory]{\includegraphics[width=\colww]{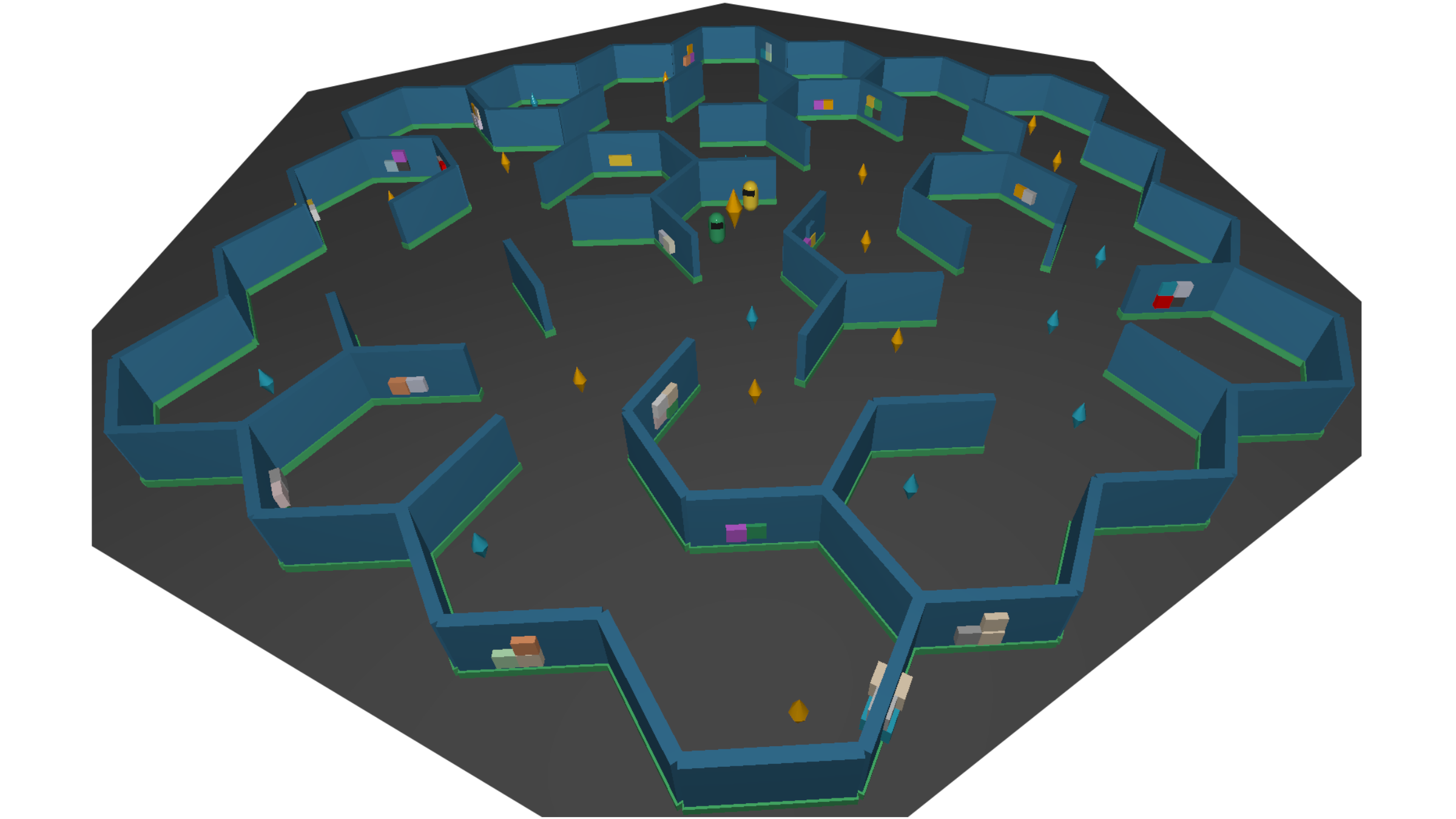}\label{fig:hexmemory}}
    \\
    \subfloat[Rearrange]{\includegraphics[width=\colww]{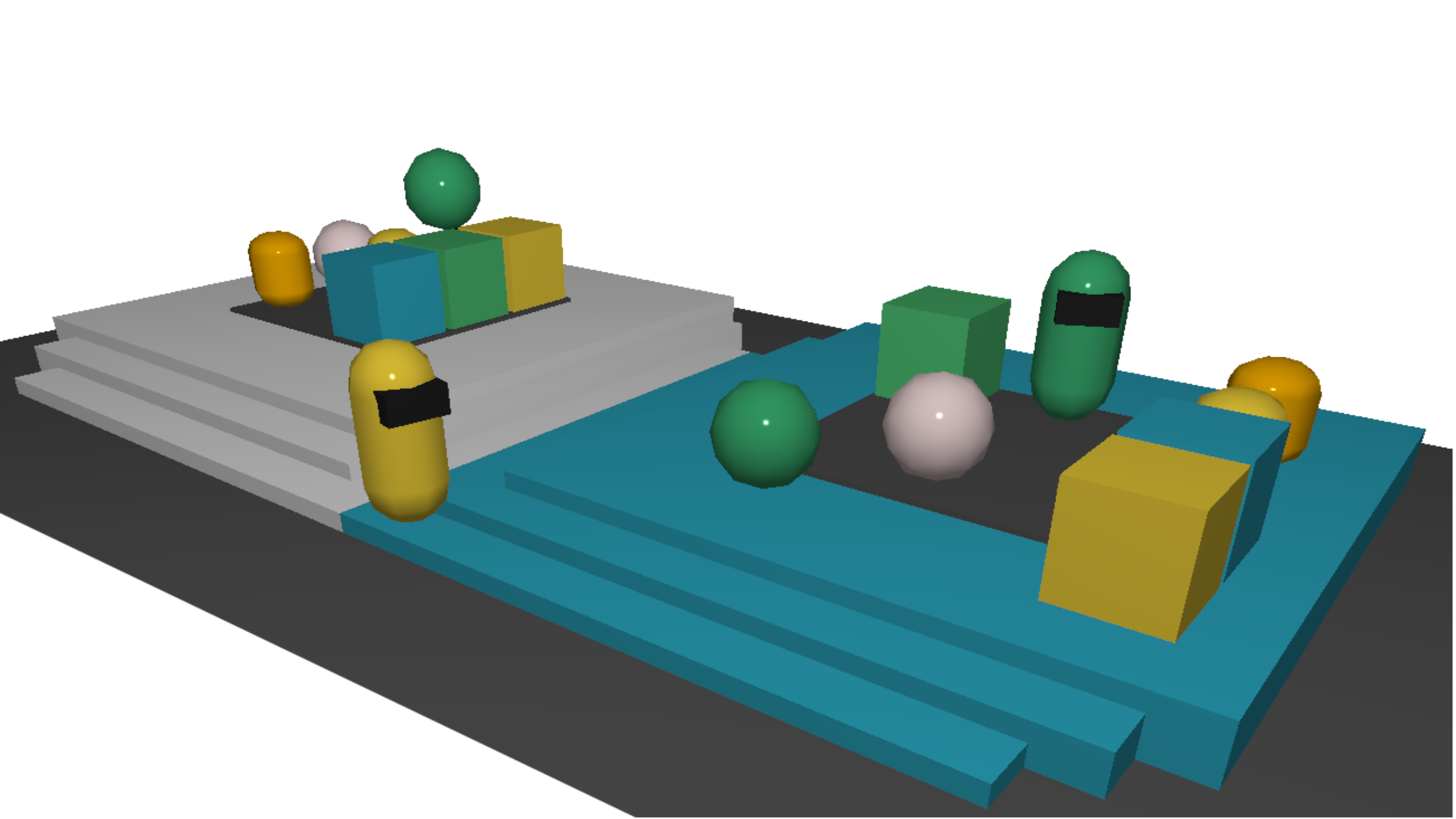}\label{fig:rearrange}}
    & \subfloat[TowerBuilding]{\includegraphics[width=\colww]{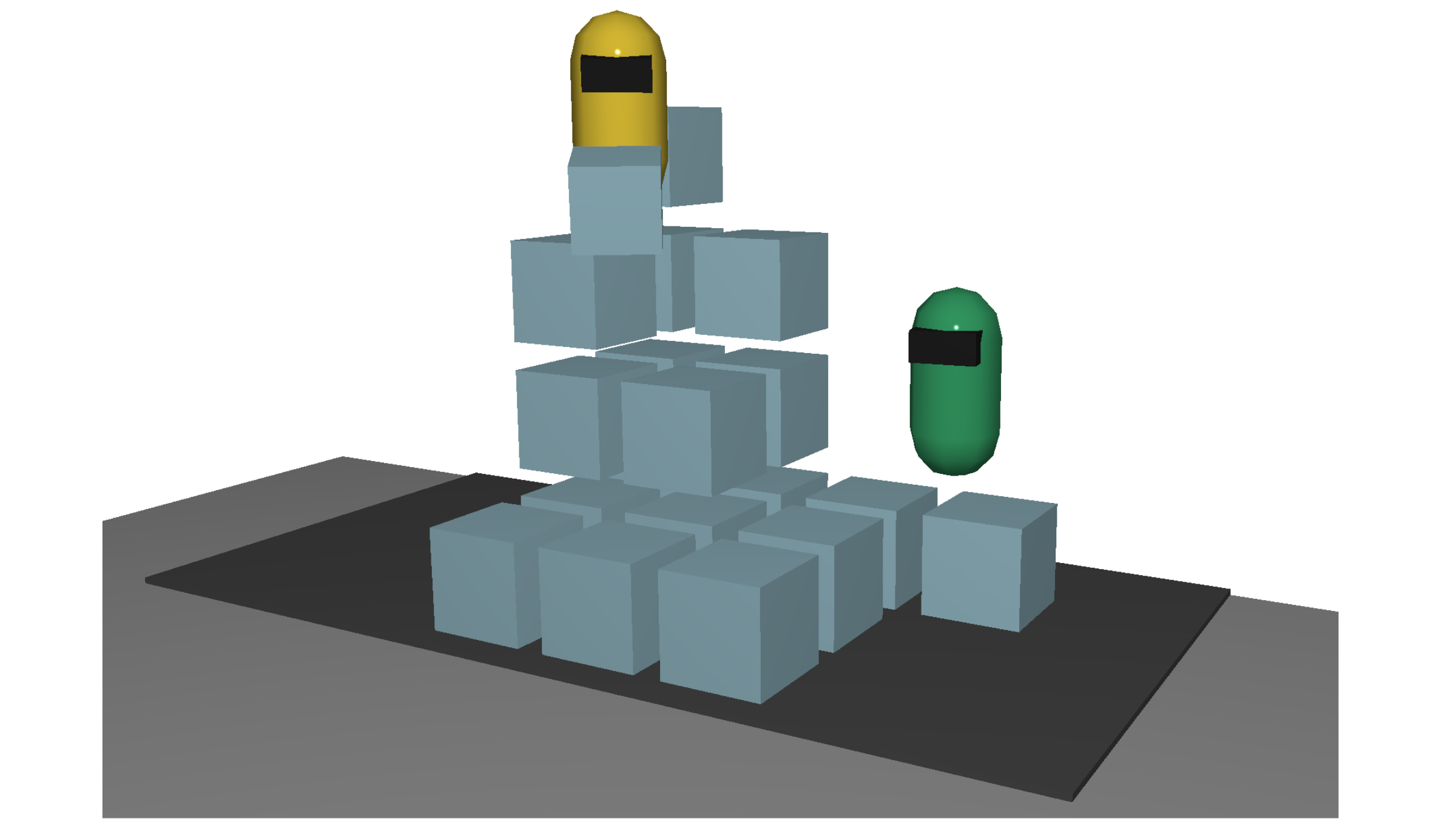}\label{fig:towerbuilding}}
    \\

\end{tabular}
\caption{ Overview of procedurally generated environments in Megaverse-8. At the beginning of each episode, the visual appearance and layout of environments are sampled randomly.}
\label{fig:megaverse_screenshots}
\end{figure*}

\csection{Megaverse-8 benchmark}

Using the Megaverse simulation platform we created a benchmark called \textit{Megaverse-8} (Figure \ref{fig:megaverse_screenshots}), designed for training and evaluation of embodied agents. The benchmark comprises eight tasks, each aiming to test different cognitive abilities of intelligent agents including exploration and navigation in 3-dimensional spaces, tool use and object manipulation, and long-term planning and memory. The benchmark is built with multi-agent support in mind, and all scenarios are suitable for teams of agents.

The environments in the benchmark are relatively simple for human-level intelligence, yet present a serious challenge for artificial agents (Section~\ref{experiments}). To score well, agents must demonstrate common-sense comprehension of the physical world, such as understanding object permanence and occlusion, and the ability to interactively manipulate objects. Creating a research platform that provides scenarios that elicit these skills is one of our motivations.

Megaverse-8 environments are procedurally generated and each task has a practically infinite number of instantiations. We randomize the parameters of the task, 3D geometry of the environment, starting positions of interactable objects and agents, \etc We randomize the visual appearance by sampling random monochrome materials. Procedural synthesis mitigates overfitting, allows us to evaluate the performance of the agents on unseen environments, and may facilitate the emergence of generalizable skills.

The following paragraphs provide high-level descriptions of the tasks in Megaverse-8. Please refer to the supplementary materials and \url{www.megaverse.info} for detailed specifications and video demonstrations.

\mypara{Megaverse ``HexExplore''} (Fig. \ref{fig:hexexplore}). Agents are placed in a randomized hexagonal maze and tasked to find a target object. The episode is considered solved when the target object is touched by the agent. This environment tests agents' episodic exploration abilities \cite{savinov2019episodic_curiosity}.

\mypara{Megaverse ``Collect''} (Fig. \ref{fig:collect}).
Agents navigate in a procedurally generated 3D landscape and collect objects. Green objects provide positive reward, while red objects generate an equal-magnitude penalty. The episode is considered solved when all positive-reward objects are collected. To generate the level geometry a 2D fractal noise texture \cite{perlin85noise} is synthesized and then interpreted as an elevation map at each location in a discretized space. This environment tests agents' skills in traversing 3D landscapes, including the ability to control the gaze direction both vertically and horizontally.

\mypara{Megaverse ``TowerBuilding''} (Fig. \ref{fig:towerbuilding}).
Agents are challenged to construct structures made of interactable boxes. Agents can pick up boxes scattered in the environment and place them in the building zone, which is marked by a distinct color. Agents receive positive reward for placing a new block in the building zone. The reward function grows as $r^{(t)} = 2^{h}$ with the height $h$ at which the block is placed, thus in order to maximize the score the agents are incentivized to construct structures with as many levels as possible. The building process is subject to realistic constraints: blocks can only be placed on top of other blocks and can only be removed if they have no blocks above them. To build non-trivial tall structures the agents need to maintain scaffolding pathways that allow them to carry the blocks to higher levels.

\mypara{Megaverse ``Sokoban''} (Fig. \ref{fig:socoban}).
Fast immersive version of the classic Sokoban puzzle, inspired by the Mujoban environment \cite{mirza2020mujoban}. At the beginning of every episode, a random puzzle with 4 boxes is sampled from the Boxoban dataset \cite{boxobanlevels}. The agents are required to push the boxes into target positions, marked green. As in classic Sokoban, some of the moves are irreversible, therefore the agents must strategically plan ahead in order to succeed.

\mypara{Megaverse ``HexMemory''} (Fig. \ref{fig:hexmemory}).
Agents are placed in a randomized hexagonal maze in front of a reference object with randomly sampled shape and material. Smaller copies of the reference object are scattered throughout the environment, alongside other objects that do not match the reference. The agent's task is to collect objects matching the reference, while avoiding other types of objects. When all matching objects are collected, the episode is terminated and the puzzle is considered solved. The scenario requires agents to memorize the visual appearance of an object and keep it in memory for long periods of time, as the reference object inevitably disappears from view as the agent navigates the maze. HexMemory challenges agents' ability to form and retain memories. This environment is inspired by \citet{doom_supercomputer}. 

\mypara{Megaverse ``Obstacles''} (Figures \ref{fig:obseasy} and \ref{fig:obshard}).
Procedurally generated 3D obstacle course presented in two versions: ObstaclesEasy and ObstaclesHard. Agents are spawned on one side of the course and are required to reach a target location on the other side. In order to get there they need to overcome different types of obstacles, such as pits, lava lakes, and high walls. Good coordination and movement is not sufficient to overcome most obstacles. For example, a wall can be too tall for the agent to jump over. Agents must use interactive objects placed in the environment to build bridges, staircases, and other artificial tools that help them accomplish the task. ObstaclesEasy and ObstaclesHard differ in both the length of the obstacle course and the difficulty of individual puzzles. ObstaclesHard is particularly difficult due to reward sparsity. As the agents are unlikely to discover the sophisticated construction behaviors by mere random exploration, the obstacle course environments can be a good test for advanced exploration strategies such as intrinsic curiosity \cite{pathakICMl17curiosity}.

\mypara{Megaverse ``Rearrange''} (Fig. \ref{fig:rearrange}).
Inspired by the classic MIT Copy Demo, this environment challenges the agents to replicate a reference structure made out of colored objects. In order to successfully complete the task the agents have to recognise and remember the object arrangement and replicate it in a designated area by rearranging interactive objects in a specific way. The task is considered solved when the reference arrangement is replicated precisely.

We expect that some of the environments, such as ``Rearrange'' and ``ObstaclesHard'', will be too challenging for present-day end-to-end learning systems. Agents not only need to discover the low-level object manipulation skills, but must also form appropriate internal representations and explore \textit{compositions} of skills in order to succeed.

\csection{Experiments}
\label{experiments}

\begin{figure*}
    \centering
    \includegraphics[width=0.95\textwidth]{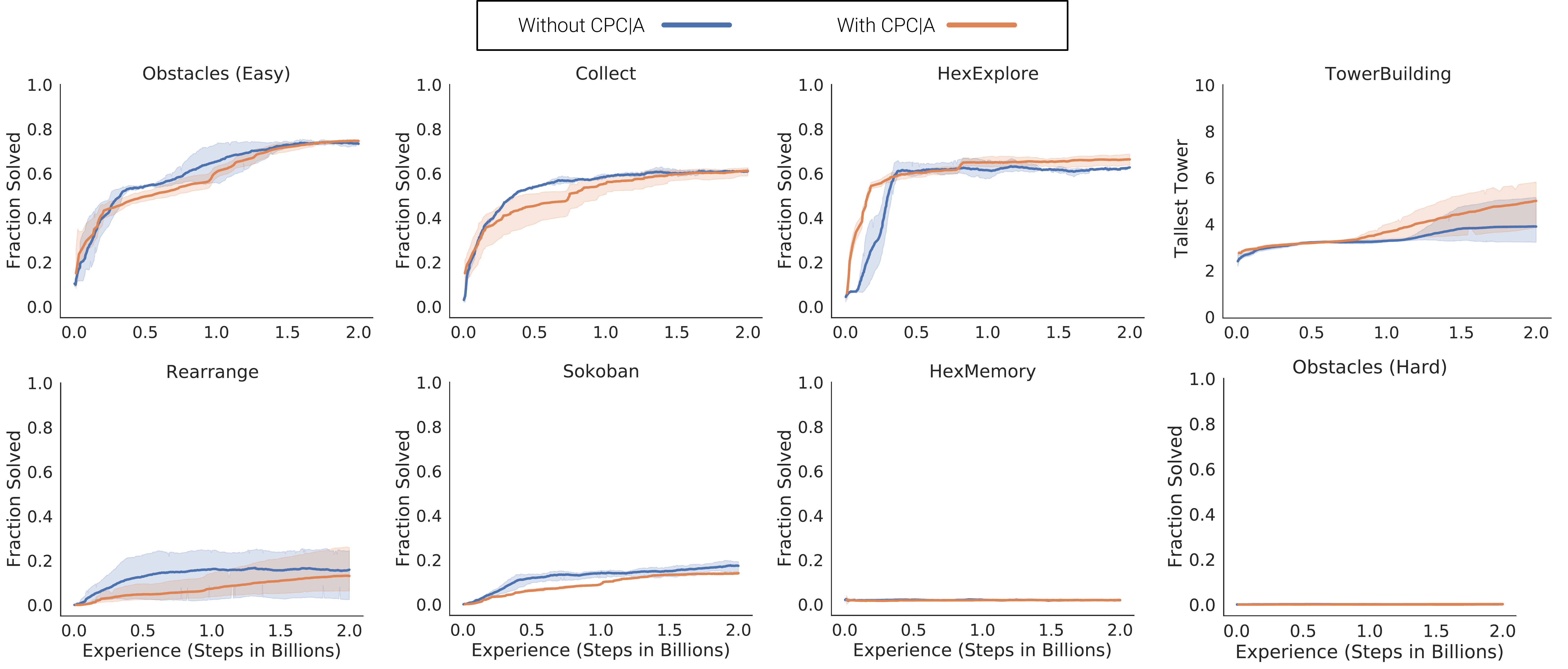}
    \caption{\textbf{Single-agent performance.} Megaverse-8 has a diversity of challenges, with environments where model-free RL is able to make progress (top four) and environments where it fails to achieve non-trivial performance (bottom four). Even for the simplest environments, there is still considerable room for improvement. The results are reported for 3 random seeds.}
    \label{fig:single-agent-results}
    \vspace{-3mm}
\end{figure*}

\begin{table}
\vspace{1mm}
\centering
\footnotesize
\begin{tabular}{l c c }

    \toprule
    
    \textbf{Environment}
    & \makecell{\textbf{Simulation} \\ \textbf{throughput}}
    & \makecell{\textbf{Training} \\ \textbf{throughput}} \\
    \midrule

    \multicolumn{3}{c}{System $\#1$ (12xCPU, 1xRTX3090)} \\
    
    \midrule
    
    Atari \scriptsize{($84\times84$ grayscale)} & 19.4k (16.8x) & 15.0k (2.8x) \\ 
    VizDoom \scriptsize{($128\times72$ RGB)} & 38.1k (8.6x) & 18.9 (2.3x) \\
    DMLab \scriptsize{($96\times72$ RGB)} & 6.1k (53.5x) & 4.6k (9.3x) \\[1mm]
    
    Megaverse \scriptsize{($128\times72$ RGB)} & \textbf{327k} & \textbf{42.7k} \\

    \midrule    
    \multicolumn{3}{c}{System $\#2$ (36xCPU, 4xRTX2080Ti)} \\
    \midrule
    
    Atari \scriptsize{($84\times84$ grayscale)} & 47.1K (18.2x) & 31.4K (2.9x) \\ 
    VizDoom \scriptsize{($128\times72$ RGB)} & 79.5K (10.8x) & 38.5K (2.3x) \\
    DMLab \scriptsize{($96\times72$ RGB)} & 12.4K (68.7x) & 7.7K (11.6x) \\[1mm]
    Megaverse \scriptsize{($128\times72$ RGB)} & \textbf{856K} & \textbf{90.1K} \\
    
    \midrule
    \multicolumn{3}{c}{System $\#3$ (48xCPU, 8xRTX2080Ti)} \\
    \midrule
    
    Atari \scriptsize{($84\times84$ grayscale)} & 53.7K (21.4x) & 34.6K (3.9x) \\ 
    VizDoom \scriptsize{($128\times72$ RGB)} & 100.1K (11.5x) & 44.7K (3x) \\
    DMLab \scriptsize{($96\times72$ RGB)} & 15.8K (72.6x) & 9.8K (13.7x) \\[1mm]
    Megaverse \scriptsize{($128\times72$ RGB)} & \textbf{1148K} & \textbf{134K} \\
    
    \bottomrule
\end{tabular}
\caption{Pure sampling and training throughput with mainstream RL simulators vs.\ Megaverse. The performance is reported in observations per second observed by the agent, i.e. after frameskip. }
\vspace{-2mm}
\label{tab:throughput}
\end{table}

\csubsection{System performance}
\label{sec:perf}

We start by benchmarking the performance of the Megaverse platform. We examine pure simulation speed, when no inference or learning are done, as well as performance of Megaverse environments as a part of a full RL training system. In order to measure training performance, we use Sample Factory \cite{petrenko2020sf}, a fast off-the-shelf RL implementation. We use three different hardware setups that are representative of systems commonly found in deep learning research labs. We compare performance of Megaverse to other fast environments used in reinforcement learning, namely Atari~\cite{ale}, VizDoom~\cite{vizdoom}, and DMLab~\cite{dmlab}.

We find that in simulation throughput, Megaverse is an order of magnitude faster than the next fastest environment, VizDoom (\mytable{tab:throughput}), while supporting considerably more complex interactions. Our platform is between 50x and 70x faster than the most comparable environment, DMLab. In end-to-end training, Megaverse is entirely bottlenecked by learning and inference throughput. However, it still enables training speeds 2-3 times faster than VizDoom and up to 14x faster than DMLab.

\begin{table}
\vspace{1mm}
\centering
\footnotesize
\begin{tabular}{c c c}
    \toprule
    \makecell{Optimized \\ geometry} & \makecell{Batched rendering \\ \cite{shacklett2021bps}} & \makecell{Simulation \\ throughput} \\
    \midrule
    \text{\sffamily X} & \text{\sffamily X} & 20.7K (10.1x) \\
    \checkmark & \text{\sffamily X} & 29.6K (7.1x) \\
    \text{\sffamily X} & \checkmark & 45.7K (4.6x) \\
    \checkmark & \checkmark & \textbf{210K} \\
    \bottomrule
\end{tabular}
\caption{Influence of optimized geometry and batched rendering optimizations on the overall sampling throughput.
Performance measured on a 10-core 1xGTX1080Ti system in Megaverse-8 ``Collect'' scenario.}
\vspace{-4mm}
\label{tab:throughput_ablation}
\end{table}

\mypara{Ablation study.}
\label{perf_ablation}
We examine the impact of two key performance optimizations in Megaverse: batched rendering (Section \ref{large_batch_rendering}) and geometry optimization (Section \ref{geometry_optimization}). The results show that both of these techniques are required to achieve high throughput (\mytable{tab:throughput_ablation}). Without geometry optimization the system would be heavily bottlenecked by physics calculations on the CPU, and without batched rendering the communication between CPU and GPU is a major bottleneck.

\csubsection{Single-agent baseline}

\mypara{Setup.} In this section we present RL training results on the Megaverse-8 benchmark. We train agents using asynchronous proximal policy optimization (PPO)~\cite{ppo} with V-trace off-policy correction~\cite{impala} using the Sample Factory implementation~\cite{petrenko2020sf}. Given the challenges of learning good representation with model-free RL from scratch, we also experiment with using Action Conditional Contrastive Predictive Coding (CPC|A)~\cite{guo2018neural} as an auxiliary loss. We train both standard PPO and a CPC|A-augmented version on $2\times10^9$ environment steps. We find that CPC|A augmentation leads to considerable performance improvements on TowerBuilding and Exploration tasks without significantly affecting other scenarios, therefore we decided to use it in all other experiments.

\mypara{Results.} We establish that the proposed benchmark has considerable diversity in task difficulty. While all tasks are far from being solved, reasonable progress can be made on four of the eight (\myfig{fig:single-agent-results}). ObstaclesEasy, Collect, and HexExplore require robust 3D navigation skills and basic object manipulation abilities. Model-free RL was able to achieve non-trivial performance in these scenarios, although none of the agents approached 100\% success.

Our agents demonstrated surprisingly high level of performance in the TowerBuilding scenario. Comparatively dense reward allowed the agents to master object stacking and consistently construct structures up to ten levels high. Video demonstrations of agent performance can be found at \url{www.megaverse.info}

Other scenarios in the benchmark have proven to be a much harder challenge. Both Rearrange and Sokoban agents improved at the beginning of training, but ultimately failed to reach satisfactory performance levels. In the Rearrange scenario, the agents learned to randomly shuffle the objects in the hope of matching the target arrangement by accident, and never learned to pay attention to the reference arrangement. In Sokoban, the agents tend to push blocks to nearby targets which happens to be sufficient to solve some of the puzzles. Full completion of the task requires long-term planning, and the agents ultimately failed to demonstrate the ability to do that.

HexMemory and ObstaclesHard turned out to be the most challenging scenarios. HexMemory is a hard credit assignment and memory challenge. Agents failed to capture the relationship between temporally distal events, such as observation of the reference object and collection of similar/dissimilar objects in the environment. This experiment shows that GRU policy networks that we used in our experiments are not sufficient for this type of task, although there is potential for other policy architectures, such as transformers \cite{parisottoSRPGJJ20transformers_rl}.

ObstaclesHard is perhaps the most challenging scenario in the Megaverse-8 benchmark. To traverse the obstacle course completely the agents need to master multiple skills and combine them in intelligent ways to overcome obstacles. Learning individual skills via a curriculum of simpler environments may be a promising research direction.

\begin{figure*}
    \centering
    \includegraphics[width=0.95\textwidth]{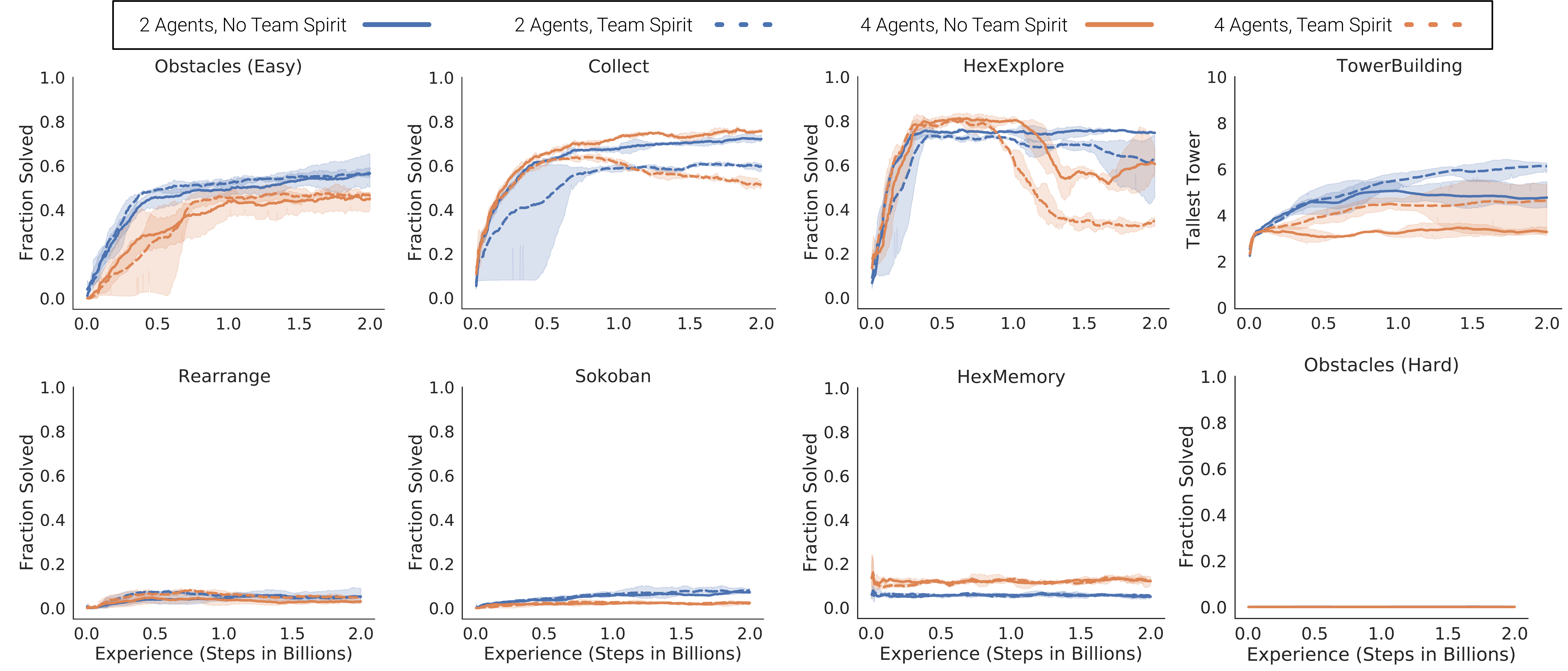}
    \caption{\textbf{Multi-agent performance.} For the majority of tasks in Megaverse-8, increasing the number of agents does not yield better results with our model-free RL framework. For all tasks except tower building, a simple Team Spirit strategy of sharing rewards has either no impact or negative impact on performance due to the increased difficulty of credit assignment. The results are reported for 3 random seeds.}
    \label{fig:multi-agent-results}
    \vspace{-3mm}
\end{figure*}

\csubsection{Multi-agent baseline}

\mypara{Setup.} We continue by examining multi-agent performance with two and four agents on our proposed benchmark. In all Megaverse-8 tasks, the agents must work together to perform well. To encourage cooperation, we experiment with Team Spirit reward shaping, inspired by OpenAI Dota 2 experiments \cite{openai2019dota}. Team Spirit modifies the credit assignment such that agents are rewarded both for their own actions and the actions of other agents in the team. Formally, the reward for agent $i$ as time $t$ is $r^{(t)}_i = (1 - \text{TeamSpirit})
\Tilde{r}^{(t)}_i + \frac{\text{TeamSpirit}}{\text{NumAgents}} \sum_j \Tilde{r}^{(t)}_j$, where $\Tilde{r}^{(t)}_i$ is the individual agent rewards before incorporating Team Spirit. This reward makes credit assignment harder, thus we gradually increase Team Spirit from 0.0 to 1.0 over the first one billion steps of training as a form of curriculum.

\mypara{Results.} For two tasks, HexExplore and Collect, we find that having more agents is beneficial (\myfig{fig:multi-agent-results}). In these tasks, relatively high score can be achieved even if agents ignore each other and focus on maximizing their own reward. This is confirmed by the fact that Team Spirit hurts performance. For HexMemory, a team of agents has a higher chance of completing the task randomly. Even though the results are better than for a single agent, teams of agents fail to make progress. In TowerBuilding, we discover that two agents perform approximately as well as one agent, and four agents consistently perform worse. The agents end up competing for rewards with each other instead of working together. In this case the addition of Team Spirit encourages cooperation and improves performance in both two- and four-agent teams.

\csection{Discussion}
\label{discussion}

We presented a new research platform \mbox{\emph{Megaverse}}, capable of achieving simulation throughput over 1,000,000 observations per second on a single node~-- unprecedented for immersive 3D simulators. Aside from the ability to simulate embodied agents tens of thousands of times faster than real time, our engine can also match the throughput of existing simulators while using only a fraction of computational resources. Our dedicated simulation platform can make large-scale RL experiments more accessible, thus accelerating progress in AI research.

Hard problems and good metrics for evaluating progress on these problems are instrumental for science. While the traditional benchmarks in Deep RL are definitely not trivial, they are getting tantalizingly close to being solved \cite{badiaPKSVGB20agent57, parisottoSRPGJJ20transformers_rl, petrenko2020sf}. We use the Megaverse platform to build \textit{Megaverse-8}, a new suite of hard challenges for embodied AI.
Complete solution of all tasks in Megaverse-8 requires the agents to master object manipulation, rearrangement, and composition of different low-level skills. We hope that solving these challenges in a robust and principled way will advance our understanding of embodied intelligence.

Extremely fast simulation provided by Megaverse can have impact beyond deep reinforcement learning. For example, contemporary derivative-free optimization methods are known for their supremacy in Mujoco-like environments \cite{SenerK20random_search}, but evaluating them in scenarios with high-dimensional observations has previously been very costly. With more than an order-of-magnitude improvement in simulation throughput, evaluation of derivative-free methods in immersive 3D environments may be feasible.

Another research direction that can leverage fast simulation is meta-learning. With highly optimized learning systems \cite{rlpyt, seedrl, petrenko2020sf}, entire training sessions in simple Megaverse environments can be completed in mere seconds, and thus can be used as a part of a larger meta-learning process. While partial learning of optimizer features and loss functions has been demonstrated \cite{bechtle2021ml3}, access to sufficiently fast training may enable the optimization of whole learners parameterized by neural networks.

Megaverse opens new possibilities in the field of multi-agent learning. Megaverse is one of the first open-source platforms that allows fast simulation of multiple agents interacting in immersive environments. The accessibility of such a platform can have important implications for studying multi-agent cooperation, autocurricula emerging from self-play \cite{hide-n-seek}, and the emergence of communication and language \cite{Mordatch18language_rl}.

{
\balance

\bibliography{bibtex}
\bibliographystyle{icml2021}
}

\onecolumn
\icmltitle{Supplementary Material}

\vskip 0.3in

\appendix

\counterwithin{figure}{section}
\counterwithin{table}{section}

\section{Megaverse: technical details}

\subsection{Interface}

Megaverse platform provides a vectorized version of the OpenAI Gym interface \citeappendix{gym} for interacting with the environments. This is a natural extension of the standard Gym interface for parallel simulators: since a single Megaverse instance simulates experience for $M$ agents in $N$ environments per step, the \textsc{step()} function accepts a vector of $N\times M$ actions, and returns a vector of $N\times M$ observations, rewards, and episode termination flags. The only difference with the original OpenAI Gym design is that Megaverse environments do not require \textsc{reset()} calls, except right after initialization. Individual simulated environments can have different episode durations, so we reset these individual environments automatically instead of relying on external code to perform resets. This has an additional performance benefit: we don't have to synthesize the last observation in the episode which is never seen by the agents. 

Although Megaverse core engine is written in C++, the high-level interface is also available through Python bindings \citeappendix{pybind11}.

\subsection{Observation and action spaces}

Megaverse provides mechanisms to configure custom observation and action spaces for individual scenarios, although all Megaverse-8 scenarios use the same unified observation and action spaces to streamline and simplify the experimentation. The observations are provided as $128 \times 72$ RGB images, and this is the only sensory input received by the agents. On top of the synthesized views of the 3D world, the observations can also contain additional information about the environment. We implement simple pixel-space GUI consisting of geometric primitives rendered in the agent's field of view. These can play the role of various bars and indicators, such as health bars, or team affiliation flags for team-based multi-agent scenarios. In Megaverse-8 scenarios we only use this GUI to notify agents about the remaining time in the episode.

Table \ref{tab:actions} describes agent's affordances. At each step the agents can independently choose walking and gaze directions, and whether they choose to jump or interact with an object. OpenAI Gym represents this action space using a tuple of discrete action spaces: \textsc{tuple(Discrete(3), Discrete(3), Discrete(3), Discrete(3), Discrete(2), Discrete(2))}. In our implementation the policy networks outputs six distinct probability vectors, which we interpret as independent categorical action distributions, although the action space can also be flattened into a single discrete action space with $324$ options.

\begin{table*}[ht]
\centering
\setlength{\tabcolsep}{3mm}
\small
\begin{tabular}{l @{\hspace{2em}} c @{\hspace{2em}} r }
    \toprule
    \textbf{Action head} & \textbf{Number of actions} & \textbf{Comment} \\
    \toprule
    Moving & 3 & no-action / forward / backward \\ 
    Strafing & 3 & no-action / left / right \\ 
    Turning & 3 & no-action / turn left / turn right \\ 
    Vertical gaze direction & 3 & no-action / look up / look down \\ 
    Jumping & 2 & no-action / jump \\ 
    Object interaction & 2 & no-action / interact \\ 
    \midrule
    Total number of possible actions & 324 & \\
    \bottomrule
\end{tabular}
\caption{Megaverse-8 action space.}
\label{tab:actions}
\end{table*}

\section{Megaverse-8}

Please refer to the project website for detailed videos demonstrating the environments: {\small{\url{www.megaverse.info}}}.

\subsection{Reward functions}

Table \ref{tab:rewards} describes the reward functions in Megaverse-8 scenarios, as seen by the learning algorithm. Besides the dense rewards that facilitate learning and exploration, Megaverse-8 environments also provide a single sparse reward (true objective) per episode that measures the real progress on the task. In all environments except TowerBuilding the true objective takes the value $+1$ when the task is successfully completed and $0$ otherwise. In the TowerBuilding scenario the true objective is to maximize the height of the structure built during the episode.

\begin{figure}
    \centering
    \includegraphics[width=0.95\textwidth]{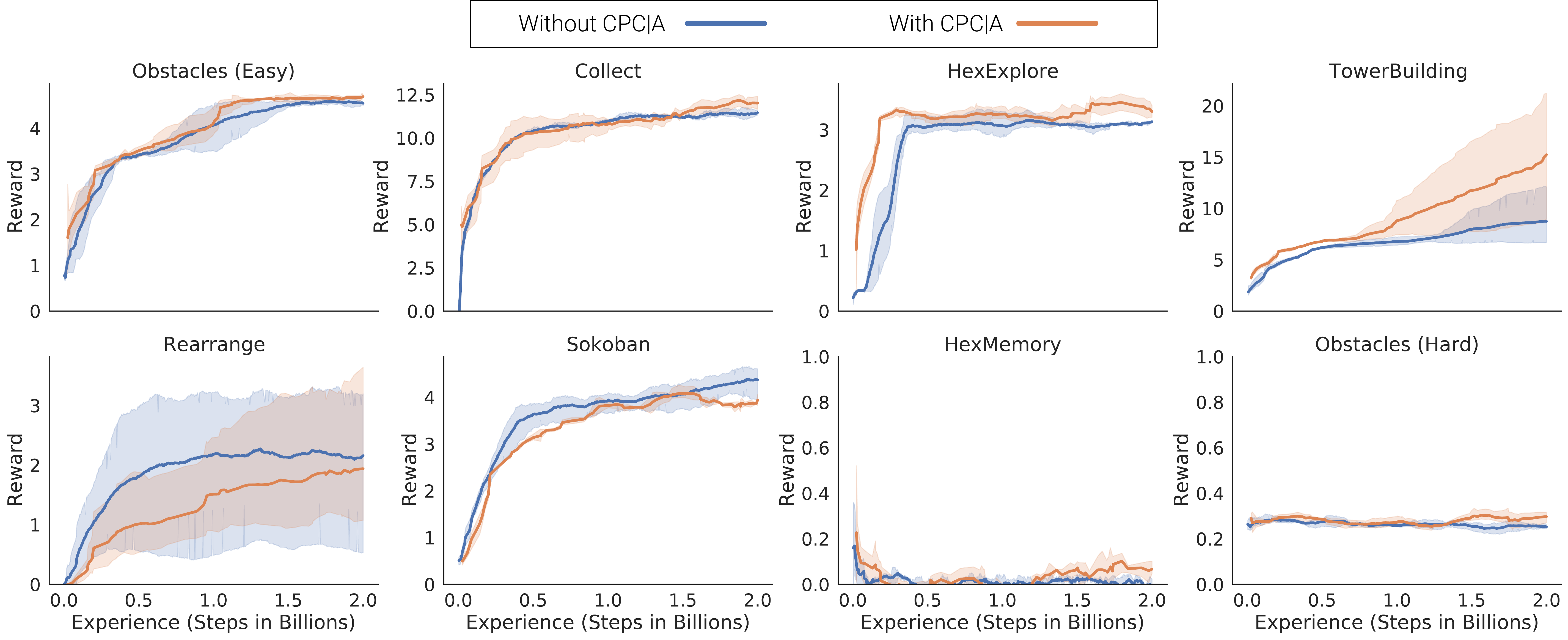}
    \caption{Total episodic reward achieved by the agents in single-agent scenarios (see reward shaping scheme in Table \ref{tab:rewards}).
    Here the results are averaged over three random seeds.}
    \label{fig:single-reward}
\end{figure}

\begin{figure}
    \centering
    \includegraphics[width=0.95\textwidth]{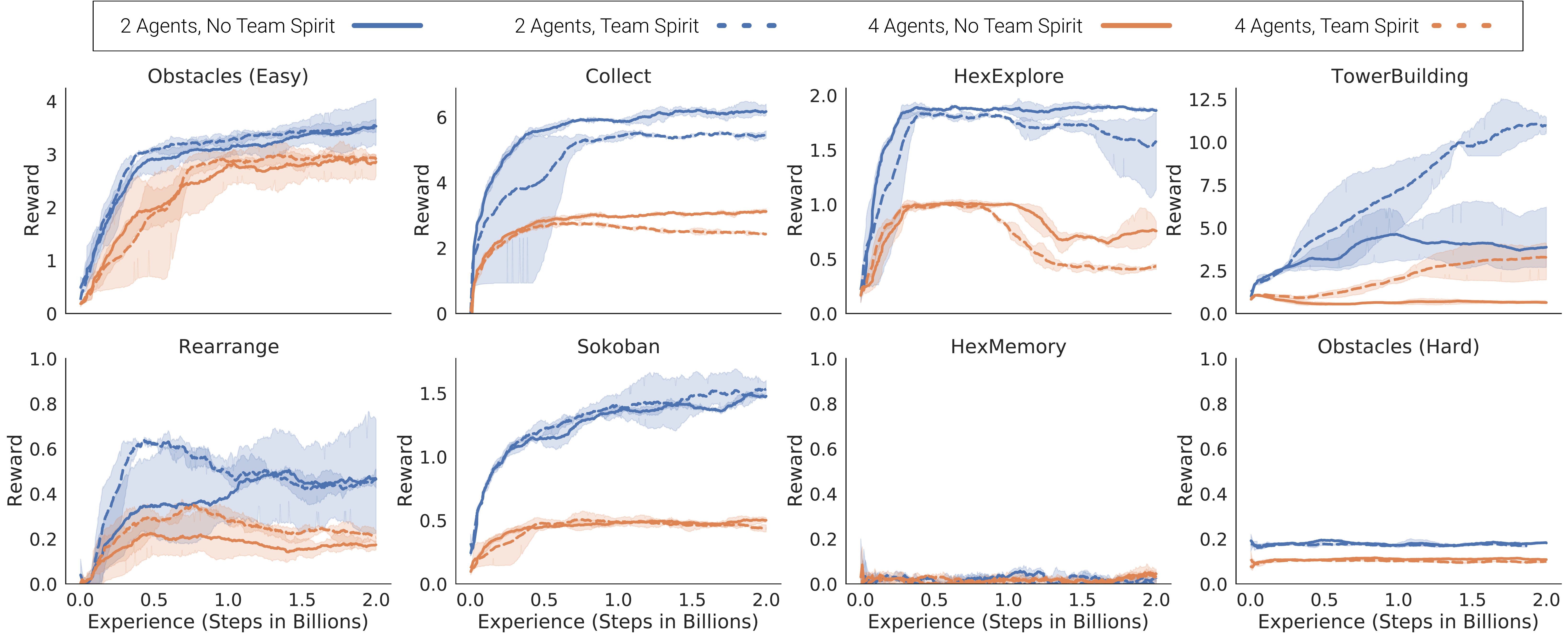}
    \caption{Total episodic reward achieved by the agents in multi-agent scenarios (see reward shaping scheme in Table \ref{tab:rewards}).
    Here the results are averaged over three random seeds.}
    \label{fig:multi-reward}
\end{figure}

In addition to task completion (true objective) results reported in the main paper, we also report dense rewards achieved by the agents in our experiments, see Figures \ref{fig:single-reward} and \ref{fig:multi-reward}.

\begin{table*}[ht]
\centering
\setlength{\tabcolsep}{3mm}
\small
\begin{tabular}{l @{\hspace{2em}} c @{\hspace{2em}} r }
    \toprule
    \textbf{Scenario} & \textbf{Dense reward} & \textbf{True objective} \\
    \toprule
    \makecell[l]{ObstaclesEasy \\ ObstaclesHard} & \makecell{$+1$ reached target location \\ $+0.5$ collected a green diamond \\ $+5$ all agents reached the target } & \makecell[r]{$+1$ (success) all agents reached the target \\ $0$ (failure) episode timed out} \\ 
    \midrule
    
    Collect & \makecell{$+1$ collecting green diamond \\ $-1$ collecting red diamond \\ $+5$ collected all green diamonds \\ $-0.5$ agent fell into the void } & \makecell[r]{$+1$ (success) collected all green diamonds \\ $0$ (failure) episode timed out} \\ 
    \midrule

    Sokoban & \makecell{$+1$ moved box onto target \\ $-1$ moved box from the target \\ $+10$ moved all boxes to targets } & \makecell[r]{$+1$ (success) moved all boxes to targets \\ $0$ (failure) episode timed out} \\ 
    \midrule
    
    HexExplore & \makecell{$+5$ found a pink diamond } & \makecell[r]{$+1$ (success) found a pink diamond \\ $0$ (failure) episode timed out} \\ 
    \midrule
    
    HexMemory & \makecell{$+1$ collected a matching object \\ $-1$ collected a non-matching object } & \makecell[r]{$+1$ (success) collected all matching objects \\ $0$ (failure) episode timed out} \\ 
    \midrule
    
    Rearrangement & \makecell{$+1$ moved object to a correct position \\ $+10$ all objects in correct positions } & \makecell[r]{$+1$ (success) all objects in correct positions \\ $0$ (failure) episode timed out} \\ 
    \midrule
    
    TowerBuilding & \makecell{$+0.1$ entered building zone with an object \\ $+0.05(h + 2^h)$ placed an object in the building zone \\ $h$ - height at which the object was placed } & \makecell[r]{$+h_{max}$ where $h_{max}$ is the max height of the tower} \\ 
    
    \bottomrule
\end{tabular}
\caption{Megaverse-8 scenarios dense rewards and final objectives.}
\label{tab:rewards}
\end{table*}

\section{Experimental details}

\subsection{Performance analysis}

Table \ref{tab:hw} provides information about hardware configuration of systems used for performance measurements. We focused on commodity hardware commonly used for deep learning experimentation.

While in the main paper we report performance figures measured only in ObstaclesHard scenario, table \ref{tab:env_performance} provides information about sampling throughput in all Megaverse-8 environments. Values represent the sampling throughput averaged over three minutes. In order to conduct the measurements we used a number of parallel Megaverse processes equal to the number of physical CPU cores with $64$ environments simulated in parallel in each process. Performance varies because different scenarios generate environments with different number of geometric primitives and interactive objects. HexMemory and HexExplore environments are based on hexagonal mazes and therefore cannot benefit from the voxel grid based optimizations that allow fast collision checking based on axis-aligned bounding boxes.

\begin{table*}[h]
\centering
\setlength{\tabcolsep}{3mm}
\small
\begin{tabular}{l @{\hspace{4em}} c @{\hspace{2em}} c @{\hspace{2em}} c}
    \toprule
    & System $\#1$ & System $\#2$ & System $\#3$ \\
    \toprule
    
    Processor & AMD Ryzen 9 3900X & Intel Xeon Gold 6154 & Intel Xeon Platinum 8280 \\ 
    Base frequency & 3.8 GHz & 3.0 GHz & 2.7 GHz  \\
    Physical cores & 12 & 36 & 48 \\ 
    Logical cores & 24 & 72 & 96 \\ 
    \midrule
    RAM & 64 GB & 256 GB & 320 GB \\ 
    \midrule
    GPUs & 1 x NVidia RTX3090 & 4 x NVidia RTX 2080Ti & 8 x NVidia RTX 2080Ti \\ 
    GPU memory & 24GB GDDR6x & 11GB GDDR6 & 11GB GDDR6 \\ 
    \midrule
    OS & Arch  (Jan 2021, Rolling) & Ubuntu 18.04 64-bit & Ubuntu 18.04 64-bit \\
    GPU drivers & NVidia 460.32.03 & NVidia 440.95.01 & NVidia 450.102.04 \\
    
    \bottomrule
\end{tabular}
\caption{Hardware configurations used for performance measurements (training and sampling performance).}
\label{tab:hw}
\end{table*}

\begin{table}
\centering
\setlength{\tabcolsep}{3mm}
\small
\begin{tabular}{l @{\hspace{2em}} r }
    \toprule
    \textbf{Scenario} & \textbf{Simulation throughput, obs/sec} \\
    \toprule
    ObstaclesEasy & $1.27\times 10^6$ \\
    ObstaclesHard & $1.15\times 10^6$ \\ 
    Collect & $8.55\times 10^5$ \\
    Sokoban & $1.16\times 10^6$ \\
    HexExplore & $6.5\times 10^5$ \\
    HexMemory & $5.9\times 10^5$ \\
    Rearrange & $1.28\times 10^6$ \\
    TowerBuilding & $1.22\times 10^6$ \\
    \bottomrule
\end{tabular}
\caption{Sampling throughput in Megaverse-8 scenarios measured on System $\#3$ (8-GPU node).}
\label{tab:env_performance}
\end{table}

\subsection{RL experiments: setup and parameters}

In all experiments in the paper we used asynchronous PPO (APPO) implementation provided by Sample Factory \citeappendix{petrenko2020sf}. Unless stated otherwise, all experiments use Action Conditional Contrastive Predictive Coding (CPC|A) \citeappendix{guo2018neural}.
For the policy network we use a small convnet model similar to VizDoom model in \citeappendix{petrenko2020sf} with a 2-layer GRU core \citeappendix{gru_Kyunghyun}.
Table \ref{tab:hyperparams} lists the learning algorithm hyperparameters.

\subsection{Additional RL experiments}

\begin{figure}
    \centering
    \includegraphics[width=0.95\textwidth]{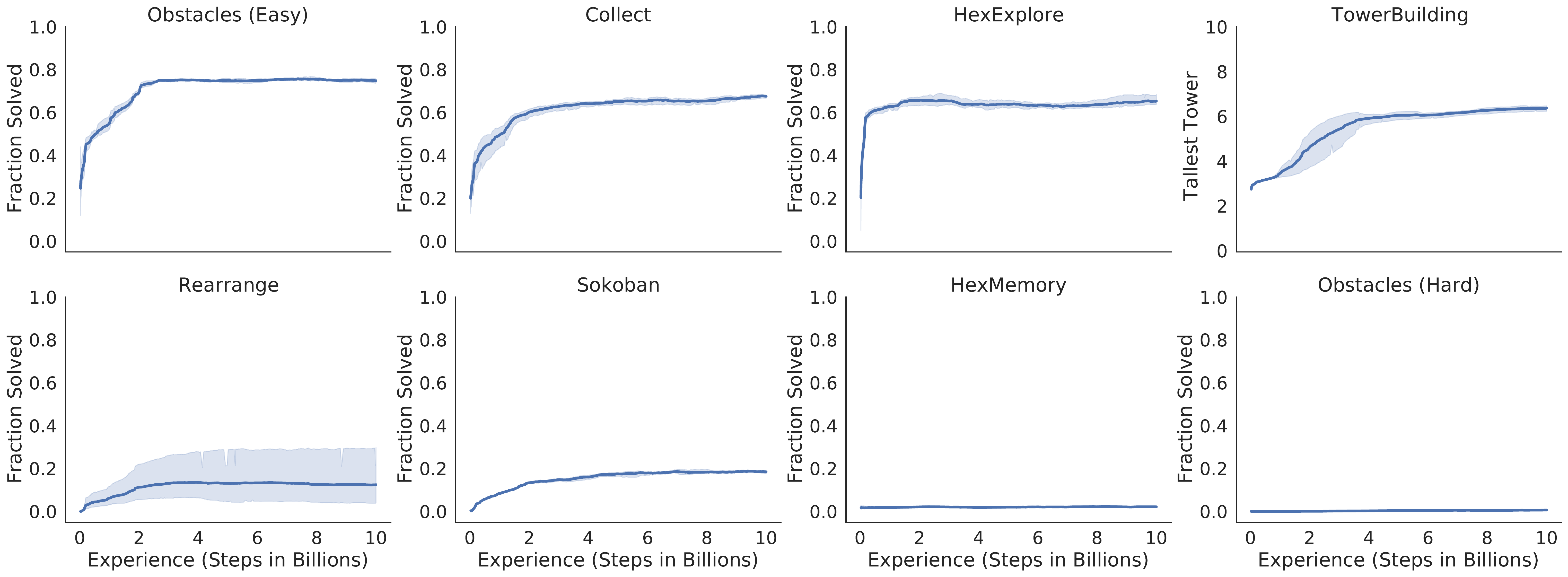}
    \includegraphics[width=0.95\textwidth]{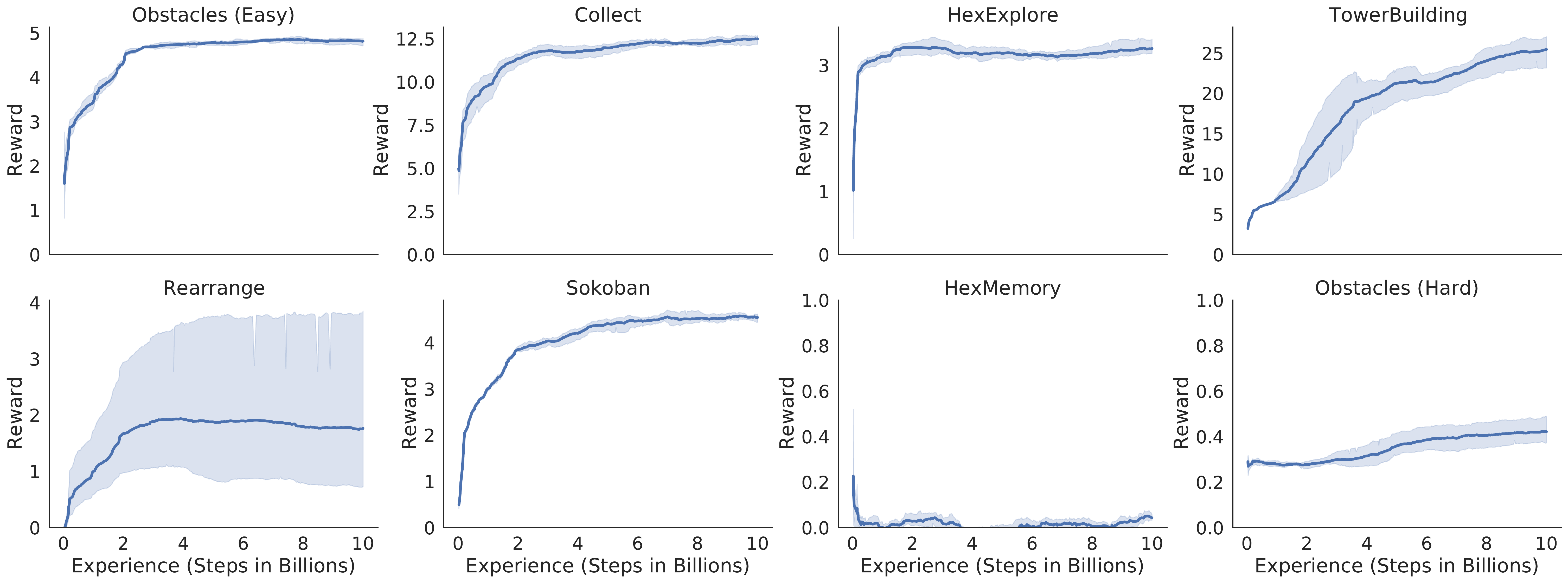}
    \caption{Single agent with CPC|A training sessions extended to $10^{10}$ frames. Both task completion (true objective) and episodic rewards are reported. Here the results are averaged over three random seeds.}
    \label{fig:10b-run}
\end{figure}

To further investigate the training performance of RL agents on Megaverse-8 tasks we conduct a series of additional experiments. First, we extended the training of the APPO+CPC|A agent in single-agent Megaverse-8 environments to $10^{10}$ environment steps (Figure \ref{fig:10b-run}). Except in TowerBuilding, we did not see a significant increase in agent's performance, which suggests that Megaverse-8 remains a challenging benchmark even in virtually unlimited sample regime (note that training for $10^{10}$ frames in Megaverse is equivalent to training for $4\times 10^{10}$ frames in DeepMind Lab or Atari due to frameskip). Instead of insufficient data, the agents are limited by their exploration abilities and cognitive capacity of relatively simple models. Thus Megaverse-8 environments can be a promising test bed for advanced exploration algorithms and policy architectures.

\begin{figure}
    \centering
    \includegraphics[width=0.95\textwidth]{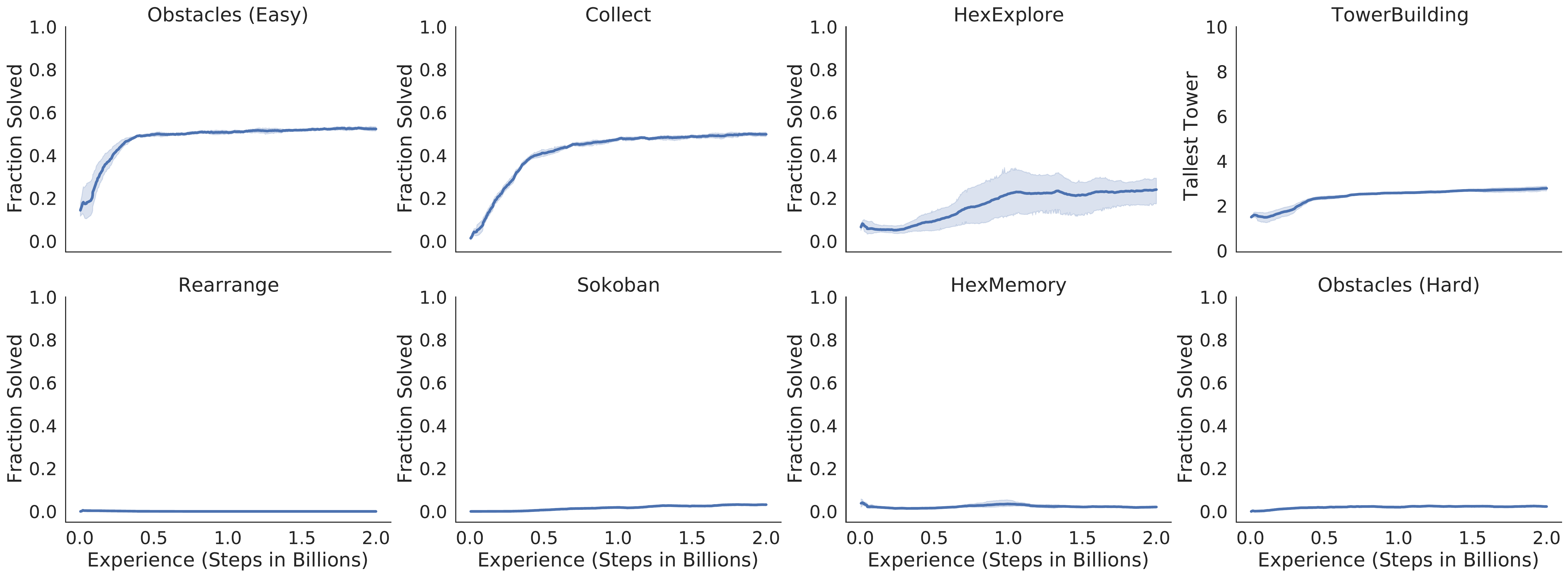}
    \includegraphics[width=0.95\textwidth]{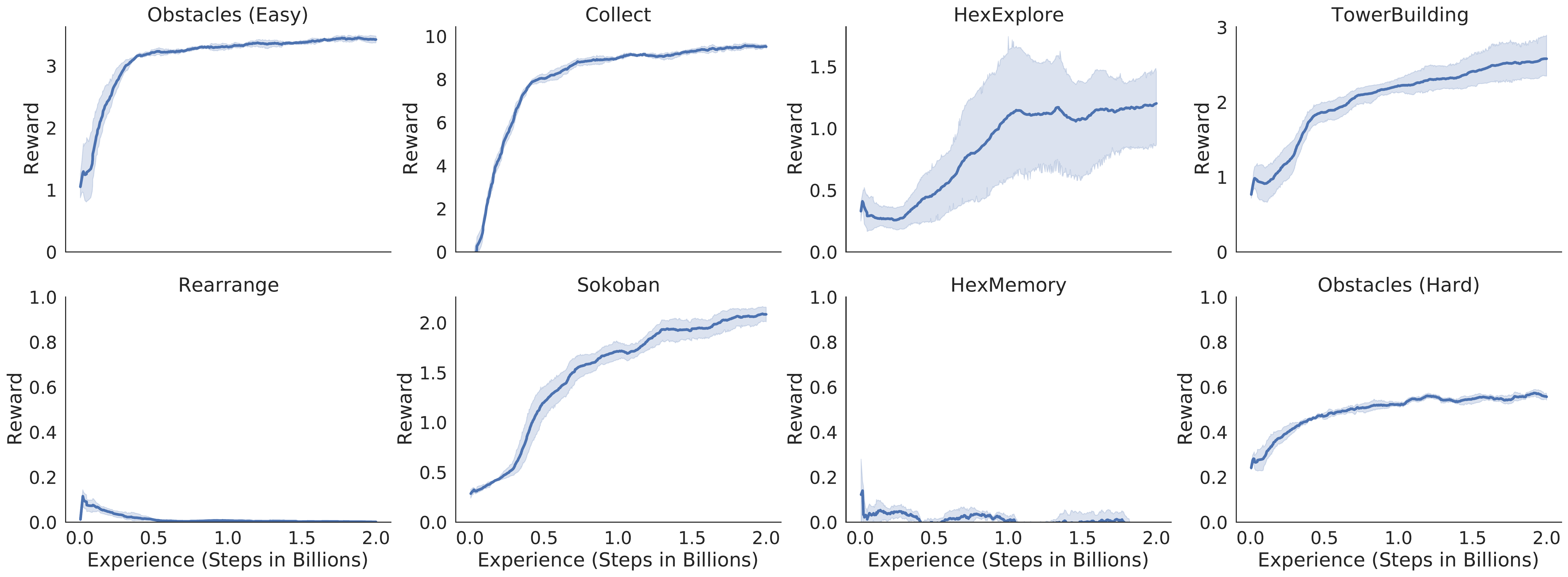}
    \caption{Performance of a single agent trained on all eight Megaverse-8 tasks. Both task completion (true objective) and episodic rewards are reported. Here the results are averaged over five random seeds.}
    \label{fig:multi-task}
\end{figure}

We also evaluated a single APPO+CPC|A agent trained on all eight Megaverse-8 environments simultaneously (Figure \ref{fig:multi-task}). The agent was trained on a total of $2\times 10^9$ frames of experience, which is equivalent to $2.5 \times 10^8$ frames on each of the environments. The results demonstrate that positive transfer can be challenging due to the diversity of Megaverse-8 tasks, although ultimately, combining experience from a diverse set of tasks and incorporating curricula can be instrumental in training capable multipurpose intelligent agents.

\begin{table}
\centering
\setlength{\tabcolsep}{3mm}
\small
\begin{tabular}{l | @{\hspace{3em}} r }
    \toprule
    Learning rate & $10^{-4}$ \\
    Action repeat (frameskip) & $1$ (no frameskip) \\
    Framestack & No \\
    Discount $\gamma$ & $0.997$ \\
    Optimizer & Adam \citeappendix{adam} \\
    Optimizer settings & $\beta_1=0.9$, $\beta_2=0.999$, $\epsilon=10^{-6}$ \\
    Gradient norm clipping & 1.0 \\
    \midrule
    Num parallel Megaverse processes & 6 \\
    Num envs simulated per process & 80 \\
    Total number of environments & 480 \\
    \midrule
    Rollout length $T$ & 32 \\
    Batch size, samples & 2048 \\
    Number of training epochs & 1 \\
    \midrule
    V-trace parameters & $\Bar{\rho}=\Bar{c}=1$  \\
    PPO clipping range & $[1.1^{-1}, 1.1]$ \\
    \midrule
    Exploration loss coefficient & $0.001$ \\
    Critic loss coefficient & $0.5$ \\
    \bottomrule
\end{tabular}
\caption{Hyperparameters in Megaverse-8 experiments.}
\label{tab:hyperparams}
\end{table}

\bibliographyappendix{bibtex}
\bibliographystyleappendix{icml2021}

\end{document}